\documentclass[10pt,journal,compsoc]{IEEEtran}
\ifCLASSOPTIONcompsoc
\usepackage[nocompress]{cite}
\else
\usepackage{cite}
\fi
\ifCLASSINFOpdf
 \usepackage[pdftex]{graphicx}
\else
\usepackage[dvips]{graphicx}
\DeclareGraphicsExtensions{.eps}
\fi
\usepackage{amsmath}
\usepackage{amssymb}
\usepackage{amsthm}

\usepackage{array}
\usepackage{url}

\usepackage[linesnumbered,ruled]{algorithm2e}

\theoremstyle{definition}

\let\oldnl\nl
\newcommand{\nonl}{\renewcommand{\nl}{\let\nl\oldnl}}

\renewcommand\thesection{\Alph{section}}

\usepackage{multirow}
\usepackage[font=scriptsize]{subfig}

\usepackage{epstopdf}
\usepackage{subfiles}
\usepackage{makecell}
\usepackage{soul}
\usepackage{color, xcolor}

\usepackage{pifont}

\usepackage{verbatim}

\soulregister\cite7 
\soulregister\eqref7
\usepackage{booktabs}
\usepackage{colortbl}
\begin{document}
	%
	\title{NavCoT: Boosting LLM-Based Vision-and-Language Navigation  via Learning Disentangled Reasoning}
	%
	%
	%
	%
	\author{Bingqian~Lin\IEEEauthorrefmark{1}, Yunshuang Nie\IEEEauthorrefmark{1}, Ziming Wei, Jiaqi Chen, Shikui Ma, Jianhua Han,\\ Hang Xu,   Xiaojun Chang, Xiaodan Liang\IEEEauthorrefmark{2}
	\IEEEcompsocitemizethanks{
	\IEEEcompsocthanksitem 
	\IEEEauthorrefmark{1}These two authors contribute equally to this work.\protect\\
	\IEEEcompsocthanksitem 
	\IEEEauthorrefmark{2}Xiaodan Liang is the corresponding author.\protect\\	 \IEEEcompsocthanksitem Bingqian Lin is with Shenzhen Campus of Sun Yat-sen University, Shenzhen, China. She is also a postdoc researcher now with Shanghai Jiao Tong University, Shanghai, China.
    \protect\\
    E-mail: linbq666@sjtu.edu.cn
    \IEEEcompsocthanksitem 
    Yunshuang Nie and Ziming Wei are with Shenzhen Campus of Sun Yat-sen University, Shenzhen, China. \protect\\
			E-mail:\{nieysh@mail2.sysu.edu.cn, weizm3@mail2.sysu.edu.cn\}
            \IEEEcompsocthanksitem Xiaodan Liang is with Shenzhen Campus of Sun Yat-sen University, Shenzhen, China, Peng Cheng Laboratory, Guangdong Key Laboratory of Big Data Analysis and Processing, Guangzhou, 510006, China.
  \protect\\
  E-mail: liangxd9@mail.sysu.edu.cn
		\IEEEcompsocthanksitem Jiaqi Chen is with the University of Hong Kong. \protect\\
 E-mail: jqchen@cs.hku.hk.
 \IEEEcompsocthanksitem Shikui Ma is with Dataa Robotics company. \protect\\
 E-mail: maskey.bj@gmail.com.\IEEEcompsocthanksitem Jianhua Han and Hang Xu are with Huawei Noah's Ark Lab. \protect\\
 E-mail: hanjianhua4@huawei.com, chromexbjxh@gmail.com.
 
 \IEEEcompsocthanksitem
 Xiaojun Chang is with the School of Information Science and Technology,
University of Science and Technology of China, Hefei 230026, China, and also
with the Department of Computer Vision, Mohamed bin Zayed University of
Artificial Intelligence (MBZUAI), Abu Dhabi, United Arab Emirates. 
 \protect\\
 Email:
cxj273@gmail.com.
	
	}
		}
	
	%
	%
	
	\markboth{IEEE TRANSACTIONS ON PATTERN ANALYSIS AND MACHINE INTELLIGENCE}%
	{Shell \MakeLowercase{\textit{et al.}}: Bare Demo of IEEEtran.cls for Computer Society Journals}
	%



	
		\IEEEtitleabstractindextext{%
		\begin{abstract}
Vision-and-Language Navigation (VLN), as a crucial research problem of Embodied AI, requires an embodied agent to navigate through complex 3D environments following natural language instructions. 
Recent research has highlighted the promising capacity of large language models (LLMs) in VLN by improving navigational reasoning accuracy and interpretability.
However, their predominant use in an offline manner usually suffers from substantial domain gap between the VLN task and the LLM training corpus. 
This paper proposes a novel strategy called \textbf{Nav}igational \textbf{C}hain-\textbf{o}f-\textbf{T}hought (NavCoT), where we fulfill parameter-efficient in-domain training to enable self-guided navigational decision, leading to a significant mitigation of the domain gap in a cost-effective manner. 
Specifically, at each timestep, the LLM is prompted to forecast the navigational chain-of-thought by:
1) acting as a world model to imagine the next observation according to the instruction, 2) selecting the candidate observation that best aligns with the imagination, and 3) determining the action based on the reasoning from the prior steps. 
In this way, the action prediction can be effectively simplified benefiting from the disentangled reasoning.
Through constructing formalized labels for training, the LLM can learn to generate desired and reasonable chain-of-thought outputs for improving the action decision. 
Experimental results across various training settings and popular VLN benchmarks (e.g., Room-to-Room (R2R), Room-across-Room (RxR), Room-for-Room (R4R)) show the significant superiority of NavCoT over the direct action prediction variants. 
Through simple parameter-efficient finetuning, our NavCoT outperforms a recent GPT4-based approach with $\sim$7\%  relative improvement on the R2R dataset. 
We believe that NavCoT will help unlock more task-adaptive and scalable LLM-based embodied agents, which are helpful for developing real-world robotics applications. 
Code is available at \url{https://github.com/expectorlin/NavCoT}.

		\end{abstract}
		
		}

	\maketitle

	\IEEEdisplaynontitleabstractindextext

	%
	\IEEEpeerreviewmaketitle



	%
		
	\IEEEraisesectionheading{\section{Introduction}\label{sec:introduction}}

\IEEEPARstart{I}{n} Vision-and-Language Navigation (VLN)~\cite{anderson2018vision,qi2020reverie,Chen2019TOUCHDOWNNL,jain2019stay,ku2020room}, an embodied agent is required to reach the target position 
following a 
language instruction. 
As a representative Embodied AI task, VLN has attracted increasing attention in recent years for its practicality and flexibility. It imposes great challenges on the embodied agent since 
successful navigation requires complex reasoning ability, e.g., 
long-term planning for following different sub-instructions and monitoring the navigation progress.

With the rapid development of the large language models (LLMs)~\cite{brown2020language,touvron2023llama,touvron2023llama2}, emerging works have attempted to introduce LLMs to solve Embodied AI tasks due to their rich real-world commonsense and powerful reasoning ability~\cite{Ahn2022DoAI,Huang2022InnerME,driess2023palme}. These works have revealed the great potential of LLMs for assisting the embodied task completion. To enable LLMs to interact with the physical world, some works have introduced the off-the-shelf vision-to-text system~\cite{li2023blip2,li2022blip} to transform the visual information into a linguistic representation. Then, the LLM can reason the action according to the textual representation of the surrounding observation. A few recent VLN works also introduce the LLM as the navigation backbone to study how LLM can improve navigation action decisions~\cite{zhou2023navgpt,long2023discuss}. However, they tend to utilize some high-cost LLMs such as GPT-4~\cite{OpenAI_2023}, which suffers from poor scalability and a large domain gap with the VLN tasks. Moreover, LLMs are required to make action decisions straightforwardly without 
guidance about how to filter noisy textual-represented visual information.

\begin{figure}[t]
\begin{centering}
\includegraphics[width=0.95\linewidth]{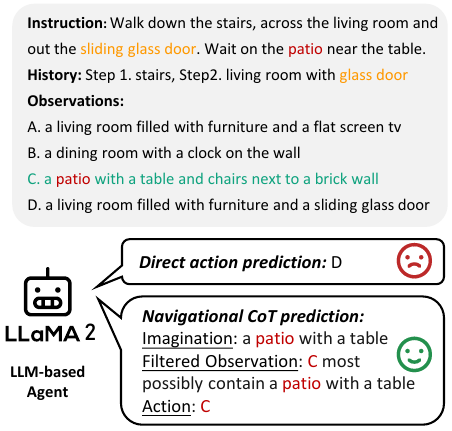}
\par\end{centering}
\caption{Comparison between direct action decision and our NavCoT. 
According to the instruction (finding {\it patio} after the {\it sliding glass door}) and history ({\it glass door}), NavCoT successfully predicts the future imagination {\it patio}, selects the observation C that best matches the imagination and determines the correct action.
}
\vspace{-0.6cm}
\label{fig:motivation}
\end{figure}

In this paper, we propose \textbf{Nav}igational \textbf{Chain}-\textbf{o}f-\textbf{T}hought (NavCoT), 
which enables LLMs to learn to perform disentangled
navigational reasoning powered by parameter-efficient in-domain training. 
Inspired by the world model theory~\cite{johnson1983mental,johnson2010mental}, when humans interact with the world, we tend to build a mental model that summarizes the surroundings we have seen before and helps us to predict the future. Then, we can make action decisions sequentially to complete different tasks based on this mental model. 
Therefore, we adapt the above process to the Chain-of-Thought (CoT)~\cite{wei2022chain} reasoning mechanism in a trainable manner.
The resulting strategy, termed {\it navigational chain-of-thought},  transforms the LLM into both a world model and a navigation reasoning agent, i.e., the LLM learns to imagine the future surroundings, filter the confusing observations based on the imagination, and then make final action decisions at each navigation timestep with the customized chain-of-thought labels.
As shown in Fig.~\ref{fig:motivation}, through NavCoT, the LLM is able to filter redundant visual information based on the imagination 
and therefore significantly simplify the action decision.



We explore various training strategies, including pretraining and finetuning, full data and low-resource settings, to study how in-domain data can contribute to the performance improvement of LLM-based VLN comprehensively. 
To facilitate training, we create formalized ground-truths to constrain the LLM to generate navigational chain-of-thoughts with a unified format.
For encouraging better instruction following, we constrain the imagination labels at each navigation timestep to be one of the mentioned objects/scenes in a given instruction, which essentially enables the construction of a task-oriented world model and  significantly simplifies the training in the meanwhile.
We adopt parameter-efficient finetuning, which can be supported by a single NVIDIA V100 GPU for two recently proposed language models (LLaMA-Adapter~\cite{zhang2023llama} and LLaMA 2~\cite{touvron2023llama2}), for improving the scalability and efficiency.


We conduct experiments on various popular VLN benchmarks, including R2R~\cite{anderson2018vision}, R4R~\cite{jain2019stay}, RxR~\cite{ku2020room}, and REVERIE~\cite{qi2020reverie}.
Experimental results show that NavCoT significantly outperforms both the direct action prediction and zero-shot inference variants, demonstrating the effectiveness of the navigational chain-of-thoughts generation in a trainable manner.
Through simple parameter-efficient finetuning, NavCoT surpasses a recent GPT4-based VLN model~\cite{zhou2023navgpt} by $\sim$7 points in both SR and SPL on R2R.

To summarize, the main contributions of this paper are: 
\begin{itemize}
\item{We introduce NavCoT, where we repurpose the LLM to be both a world model and a navigation reasoning agent in a trainable manner to simplify the action decision process and improve interpretability.}
\item{We adopt parameter-efficient in-domain training for adapting LLMs to the VLN task in a cost-effective way, making a solid step towards developing scalable LLM-based VLN methods.}
\item{Experimental results show the superiority of NavCoT over high-cost LLM-based approaches and direct action prediction variants on multiple VLN datasets. Through explicit reasoning generation, NavCoT also exhibits much better explainability than traditional cross-modal based VLN models.}
\end{itemize}

\section{Related Work}

\subsection{Vision-Language Navigation}
VLN has received great attention and many works have been proposed in the past few years. 
Early approaches mainly focus on exploring data augmentation techniques~\cite{tan2019learning, fried2018speaker,liu2021vision,Fu2019CounterfactualVN}, introducing learning mechanisms~\cite{liang2022contrastive,zhu2020vision,lin2021adversarial,lin2022adapt}, and designing model architectures~\cite{wang2019reinforced,ma2019self, deng2020evolving,qi2020Object} to alleviate data scarcity and enhance the navigation performance. 
To further improve the generalization to unseen environments, pretraining-based approaches~\cite{hong2021vln,Chen2021HistoryAM,Chen2022ThinkGA,Qiao2022HOPHA,Guhur2021AirbertIP,an2022bevbert,wang2023scaling} 
have been widely developed
in the VLN field. 
However, adapting to realistic application scenarios that require rich commonsense knowledge is still quite challenging for existing VLN agents.
Furthermore, previous methods usually lack enough interpretability in action decisions. 
To address the above issues, a few recent works introduce LLMs with great knowledge storage as the navigation backbone in a zero-shot manner~\cite{zhou2023navgpt,long2023discuss,chen2024mapgpt,chen2024affordances}. 
NavGPT~\cite{zhou2023navgpt} constructs a purely LLM-based navigation agent that receives textual represented observations and navigation histories to make action decisions. 
DiscussNav~\cite{long2023discuss} incorporates large models with unique capabilities to act as domain experts and asks the agent to actively discuss with these experts to make decisions. MapGPT~\cite{chen2024mapgpt} uses an online constructed language-formalized map to encourage the agent for global exploration and adaptive planning. 
Nevertheless, the severe domain gap and the dependency on high-cost LLMs significantly harm navigation performance and scalability.
NaviLLM~\cite{zheng2024towards} constructs a trainable LLM-based agent to make LLMs to better adapt to VLN tasks. However, it directly maps navigation inputs to action predictions without the reasoning output.

This paper proposes a new LLM-based VLN approach called NavCoT,
where we conduct parameter-efficient in-domain training for teaching LLMs to 
perform disentangled navigational reasoning accurately for facilitating action decisions. Our NavCoT significantly improves the scalability of LLM-based VLN agents and narrows the domain gap between LLM's training corpus and the VLN task. 
Through navigational chain-of-thought reasoning generation in a trainable manner, our NavCoT improves the accuracy of both navigation reasoning and action decisions.

\subsection{LLMs for Embodied AI}
Introducing large language models (LLMs) into Embodied AI tasks has gained widespread interest recently. Benefiting from training on the ultra-large-scale corpus, LLMs exhibit the brilliant ability of planning, reasoning, and reflection to assist embodied task completion~\cite{Ahn2022DoAI,Huang2022InnerME,Yao2022ReActSR,shahlm,schumann-2023-velma,wang2023voyager}. SayCan~\cite{Ahn2022DoAI} combines LLMs with affordance functions to produce feasible plans for completing household tasks. Inner Monologue~\cite{Huang2022InnerME} makes further improvements on~\cite{Ahn2022DoAI} by injecting feedbacks from the environment. Although these approaches enable LLMs to interact with specific environments in different tasks, their offline use of LLMs inevitably brings noise. Recent methods have employed in-domain training to better adapt LLMs to embodied tasks~\cite{mu2023embodiedgpt,yang2023octopus,brohan2023rt}. For example, EmbodiedGPT~\cite{mu2023embodiedgpt} crafts a large-scale embodied planning dataset and adapts the LLM to it by introducing an additional embodied-former. In our work, we adopt a parameter-efficient training scheme and utilize the available VLN training data to construct the navigational chain-of-thought labels, which effectively adapts the LLM to VLN tasks at a much lower cost.

\subsection{Chain-of-Thought Prompting}
Chain-of-Thought (CoT) prompting, firstly proposed in~\cite{wei2022chain}, is a powerful in-context learning technique to elicit multi-step reasoning abilities of LLMs. By elaborating intermediate reasoning steps to form the CoT rather than generating the answer only in the prompt, LLMs can learn to generate the output accordingly for the specific task and therefore improve the reasoning accuracy. Following~\cite{wei2022chain}, different works improve standard CoTs through self-consistency~\cite{wang2022self}, least-to-most prompting~\cite{zhou2022least}, boostrapping~\cite{zelikman2022star}, tree-of-thought prompting~\cite{yao2023tree,long2023large}, {\it etc}. However, most of them prompt LLMs to produce CoTs in an offline and unconstrained manner. 

In this work, we introduce the theory of the world model into the CoT mechanism in a trainable way and constrain the LLM to produce CoT outputs with the unified format by collecting formalized ground-truths. As a result, the LLM can learn to produce self-guided navigational reasoning, and the training process can be greatly simplified.


\section{Preliminaries}
\label{Preliminaries}
\subsection{Problem Setup}
VLN requires an agent to follow a language instruction $I$ to navigate from a start viewpoint to the target viewpoint. 
At timestep $t$, the agent receives a panoramic observation $O_{t}$ containing $K$ single-view observations $O_{t,k}$, i.e., $O_{t}=\{O_{t,k}\}_{k=1}^{K}$.
There are $N$ navigable views 
among $K$ views. 
The 
navigable
views 
and a stop token $[stop]$ form the action space, 
from which the agent chooses one as the action prediction $a_{t}$.
Actions before step $t$ are viewed as the navigation history, which is denoted as $H_{t}=\{a_{0},...,a_{t-1}\}$.
A navigation trajectory is successful when the agent stops within 3m of the target viewpoint.

\subsection{Large Language Models (LLMs)}
Applying LLMs to non-linguistic embodied tasks has received more and more attention recently. We roughly divide these methods into two categories: the first one is to employ closed-source LLMs such as GPT-4~\cite{OpenAI_2023} in an offline way~\cite{zhou2023navgpt,shahlm,Ahn2022DoAI}, which may suffer from poor scalability and severe domain gap. The second one is to introduce smaller open-source LLMs 
which can be deployed and trained locally~\cite{mu2023embodiedgpt}. Our method lies in the latter, and we adopt two open-source LLMs, LLaMA-Adapter~\cite{zhang2023llama} and LLaMA 2~\cite{touvron2023llama2}, as the navigation backbones.

LLaMA-Adapter~\cite{zhang2023llama} is a lightweight adaption method that finetunes LLaMA 1~\cite{touvron2023llama} efficiently with less time and parameters. The core idea is to introduce learnable adaption prompts and a zero-initialized attention mechanism with zero gating. LLaMA-Adapter can generate comparable responses to Alpaca 7B~\cite{alpaca} trained with fully fine-tuned parameters.
LLaMA 2~\cite{touvron2023llama2} is an updated version of LLaMA 1~\cite{touvron2023llama}. It is trained on 2 trillion tokens and with twice the context length of the LLaMA 1. LLaMA 2 contains variants with different parameters scales, such as 7B, 13B, and 70B. We adopt a bias tuning strategy~\cite{gao2023llama} for implementing parameter-efficient fine-tuning on LLaMA 2 7B.

\begin{figure*}[t]
\begin{centering}
\includegraphics[width=0.98\linewidth]{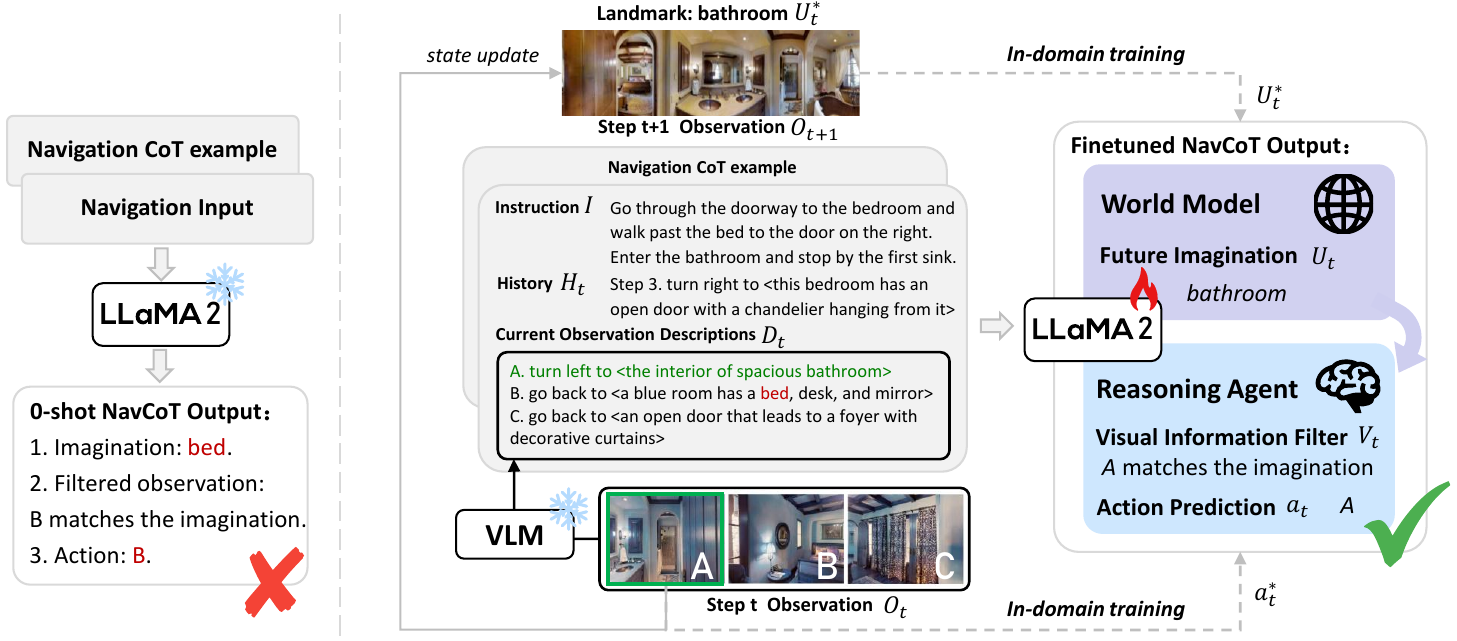}
\par\end{centering}
\caption{Overview of NavCoT. At timestep $t$, we employ a VLM to translate the observation information into textual description. Then, the LLM is prompted with the example and the textual represented navigation input to produce the navigational chain-of-thought. We conduct in-domain training to enable the LLM to learn to generate reasonable navigational reasoning for action decisions.
}\label{fig:overview}
\vspace{-0.4cm}
\end{figure*}

\section{Method}
The overview of NavCoT is presented in Fig.~\ref{fig:overview}. 
At each timestep $t$, we adopt a vision-to-text system to convert the surrounding observation into linguistic representations (Sec.~\ref{Vision-to-Text-System}). Then, we prompt the LLM with the in-context example and textual represented navigation input to produce the disentangled navigational reasoning, i.e., navigational chain-of-thought (CoT)  (Sec.~\ref{Design of Chain-of-Thought Prompt}). We collect customized reasoning labels (Sec.~\ref{Ground-Truth Collection}) and adopt various training settings such as imitation-learning based finetuning and task-decomposed pretraining (Sec.~\ref{Training and Inference}) for implementing parameter-efficient in-domain training. As a result, the LLM learns to generate correct disentangled reasoning with constrained formats for improving the action decision accuracy.


\subsection{Vision-to-Text System}
\label{Vision-to-Text-System}
The observation $O_{t,n}$ at each timestep $t$ contains an RGB image $B_{t,n}$ and the direction information $A_{t,n}=\{\psi_{t,n},\theta_{t,n}\}$, where $\psi_{t,n}$ and $\theta_{t,n}$ represent the heading and elevation, respectively. Therefore, the vision-to-text system translates both the vision information in $B_{t,n}$ and the direction information $A_{t,n}$ into textual descriptions and feeds them into the LLM for action decisions. We use an image captioning model BLIP~\cite{li2022blip}, denoted as $F_{v}$, to translate the vision information in $B_{t,n}$ to a caption ${D}^{v}_{t,n}$:
\begin{equation}
{D}^{v}_{t,n} = F_{v}(B_{t,n}).
\end{equation}
We map the direction information $A_{t,n}$ into the textual represented direction space containing six basic directions such as ``turn left'' and ``go up'', following the direction mapping rules in VLN~\cite{anderson2018vision}.
Denote the mapped direction information as ${D}^{a}_{t,n}$, the final textual description ${D}_{t,n}$ for each observation $O_{t,n}$ is obtained by: 
\begin{equation}
{D}_{t,n} = \mathrm{cat} ({D}^{a}_{t,n},{D}^{v}_{t,n}),
\end{equation}
where $\mathrm{cat}(\cdot)$ denotes the string concatenation. 
For convenience, we add the alphabetical represented label for each observation to convert it into an action option, as in Fig.~\ref{fig:overview}.

\subsection{Navigational Chain-of-Thought Prompt}
\label{Design of Chain-of-Thought Prompt}
Proper design of the intermediate reasoning steps is crucial in designing the chain-of-thought (CoT) prompt, which may greatly affect the performance of LLM's predictions. In this work, we aim to empower the LLM to generate two significant intermediate reasoning steps for guiding the navigation action predictions, inspired by the world model theory~\cite{johnson1983mental,johnson2010mental}. Specifically,  we formalize the navigational chain-of-thoughts to contain three important steps, i.e., Future Imagination (FI), Visual Information Filter (VIF), and Action Prediction (AP). FI enables the agent to predict possible objects/scenes it may encounter with the guidance of the instruction, which helps the agent capture the environmental dynamics and monitor the progress of the navigation. VIF explicitly connects the generated imagination with the subsequent action prediction. 

As shown in Fig.~\ref{fig:overview}, at each timestep $t$, the LLM receives the prompt consisting of a chain-of-thought reasoning example and a query navigation input. The reasoning example serves as a reference to guide the LLM to generate the desired format of reasoning based on the given navigation input. The navigation input at timestep $t$ consists of the instruction $I$, the textual described observation $D_{t}$ obtained in Sec.~\ref{Vision-to-Text-System}, i.e., $D_{t}$ = $\{D_{t,n}\}^{N}_{n=1}$ (N is the number of navigable views), and the navigation history $H_{t}$. 
With the navigation input, we prompt the LLM to generate the constrained reasoning format for the FI, VIF and AP steps.

\noindent\textbf{Future Imagination (FI).}
In FI, we want LLM to generate the imagination about the next observation, which can be an object or a scene. 
Denote the imagination generated by LLM as $U_{t}$, the desired output format for FI is: 

\texttt{Imagination: $U_{t}$}.

\noindent\textbf{Visual Information Filter (VIF).}  
After generating the imagination in FI, we introduce a further reasoning step, VIF, to force LLM to explicitly select the observation that best matches the imagination from the redundant observation information.
Denote the option of observation that LLM predicts to align the imagination $U_{t}$ best as $V_{t}$. 
The desired output format for VIF is:

\texttt{Filtered observation: $V_{t}$ matches the imagination.}





\noindent\textbf{Action Prediction (AP).} 
By summarizing the reasoning in FI and VIF, the LLM can make the final action prediction.  
Denote the option of action that LLM predicts as $a_{t}$, 
we define the output format for AP as:

\texttt{Action: $a_{t}$.}

We 
give an in-context 
example to the LLM as follows:

\texttt{Input: Instruction: Walk towards the mirror and walk through the open door. Observation: [A. stop,  B. go forward to <a bedroom with a bed>, C. turn right to <an open door leading to a hallway>]. History: Step 1. go forward to <a wall with a mirror>.}

\texttt{Output: Imagination: open door. Filtered observation: C matches the imagination. Action: C.}


Through this example, the LLM can know the desired reasoning format and principle, e.g., the navigation history indicates ``mirror'' and the resulting imagination is ``open door''.
Based on the given example, we ask the LLM to generate the desired reasoning by giving the following prompt:

\texttt{Input:~Instruction:~\{$I$\} Observation: \{$D_{t}$\} History:~\{$H_{t}$\}}


\texttt{Output:}

\begin{figure*}[t]
\begin{centering}
\includegraphics[width=0.95\linewidth]{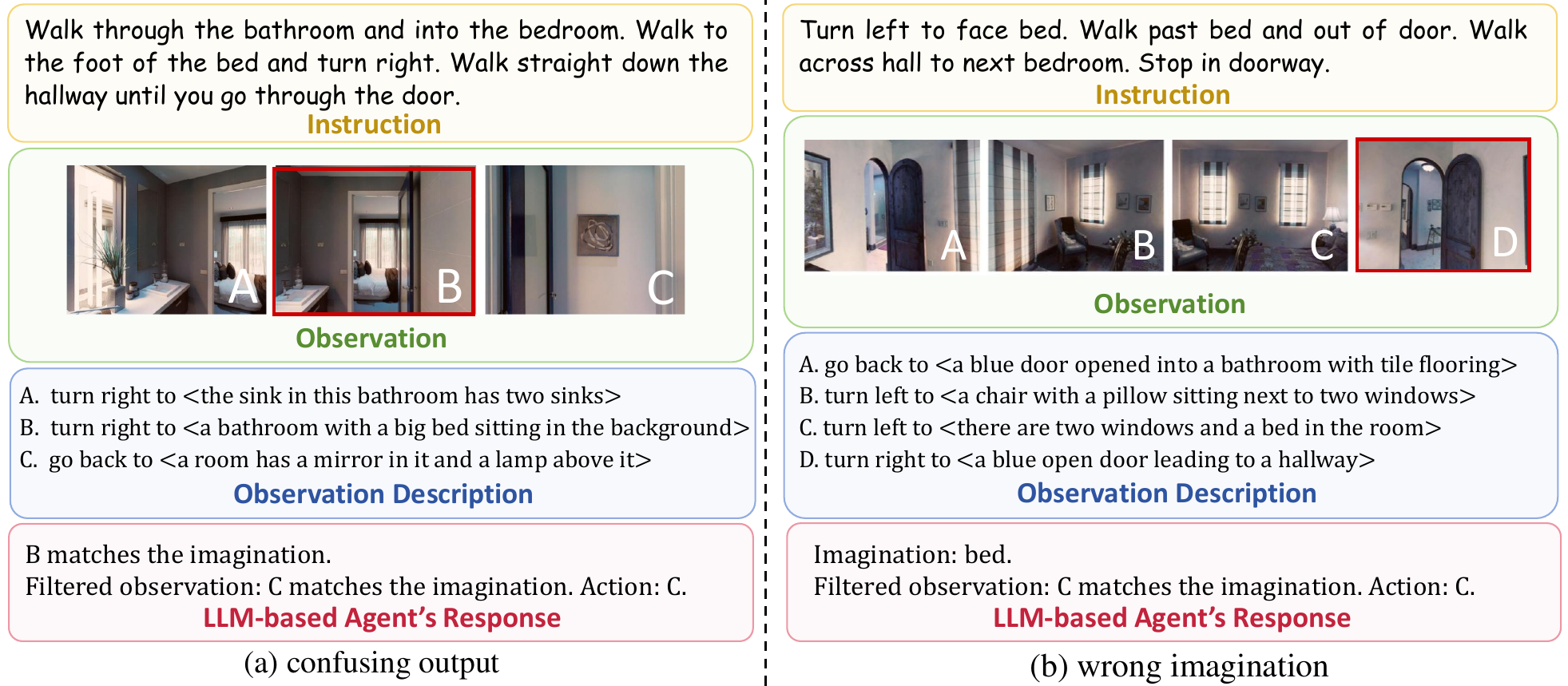}
\par\end{centering}
\vspace{-0.2cm}
\caption{Failure cases of LLM output in the zero-shot manner. The ground-truth actions are denoted by red boxes.
}
\vspace{-0.4cm}
\label{fig:llm output}
\end{figure*}

\subsection{Reasoning Ground-Truth Collection} \label{Ground-Truth Collection}
As shown in Fig.~\ref{fig:llm output}, due to the uncertainty of the LLM's output and the complexity of the VLN task, it is hard for the LLM to generate the multi-step reasoning accurately for action decisions in a zero-shot manner. 
For example, in Fig.~\ref{fig:llm output}(a), the LLM produces confusing reasoning output for the visual information filter. In Fig.~\ref{fig:llm output}(b), the LLM generates the wrong imagination ``bed'' that would mislead itself to choose the wrong action.
To effectively adapt the LLM for complex navigation decisions in the VLN task, we collect the ground-truth of the navigational chain-of-thought based on the available VLN data for implementing in-domain training.

We firstly collect the ground-truth imagination ${U}^{*}_{t}$ for the reasoning task FI. 
Ideally, the ground-truth 
imagination should be consistent with the object/scene appearing in the following observation, which corresponds to the ground-truth action. Moreover, to enable better cross-modal alignment between the observation and the instruction for action decisions, the imagination is desired to be one of the mentioned objects/scenes in the given instruction. To this end, 
we resort to an LLM to extract the mentioned objects/scenes in the instruction and a cross-modal large model CLIP~\cite{yu2021learning} to collect the  ground-truth imagination ${U}^{*}_{t}$ at different timesteps. Concretely, for a given instruction $I$, we provide the  prompt with the customized task example to ask the LLM to extract the mentioned objects/scenes from $I$. 
Denote the extracted 
landmark list from $I$ as $U^{la}=\{U^{la}_{k}\}_{k=1}^{M}$, where $M$ is the number of objects/scenes mentioned in the instruction $I$. For the ground-truth observation $B^{*}_{t}$ at each timestep $t$, we calculate the similarity between 
$B^{*}_{t}$
and each 
landmark $U^{la}_{k}$ in the list $U^{la}$, 
and take
the one with the highest similarity as the ground-truth imagination label ${U}^{*}_{t}$:
\begin{equation}
{U}^{*}_{t} = \mathop{\mathrm{argmax}}\limits_{U^{la}_{k}}\mathrm{Sim}(F^{t}_{\mathrm{CLIP}}(U^{la}_{k}), F^{v}_{\mathrm{CLIP}}(B^{*}_{t})),
\end{equation}
where $F^{t}_{\mathrm{CLIP}}$ and $F^{v}_{\mathrm{CLIP}}$ represent the text encoder and the image encoder of CLIP, respectively. 

Since the reasoning task VIF aims to 
find aligned observation with the instruction for action decision, we 
set the ground-truth label of filtered observation to be consistent with the 
option of ground-truth action $a^{*}_{t}$.
As a result, the ground truth of the navigational chain-of-thought, denoted as $\mathrm{CoT}^{*}_{t}$, for 
instruction $I$ at timestep $t$ is defined by:

\texttt{Imagination:~${U}^{*}_{t}$.~Filtered~observation: $a^{*}_{t}$ matches the imagination. Action:~$a^{*}_{t}$.}

\subsection{In-domain Chain-of-Thought Training}
Pretraining and finetuning are two widely used techniques for training VLN agents~\cite{hao2020towards,Chen2021HistoryAM,Chen2022ThinkGA}. Therefore, we also conduct both pretraining and finetuning with VLN data for fulfilling our in-domain training scheme. Specifically, we decompose each reasoning procedure in the navigational chain-of-thought (CoT) as a separate task to implement multi-task pretraining. As a result, the accuracy of each reasoning procedure can be effectively promoted for facilitating the subsequent finetuning. After pretraining, we conduct finetuning with the imitation learning strategy~\cite{anderson2018vision} to ask the LLM to produce complete navigation CoT reasoning, which further adapts LLM to generate self-guided navigational decisions sequentially for successful navigation. 

\label{Training and Inference}


\noindent\textbf{Pretraining.}
In NavCoT, 
we set each of the three navigational reasoning tasks defined in Sec.~\ref{Design of Chain-of-Thought Prompt} as a pretraining 
task and create corresponding instruction-following dataset. 
The pretraining objective $\mathcal{L}_{p}$ is defined as follows:
\begin{equation}
\mathcal{L}_{\mathrm{FI}} = -U^{*}\mathrm{log}(p_{\mathrm{LLM}}(U|I,H,D)),
\end{equation}
\begin{equation}
\mathcal{L}_{\mathrm{VIF}} = -V^{*}\mathrm{log}(p_{\mathrm{LLM}}(V|I,H,D)),
\end{equation}
\begin{equation}
\mathcal{L}_{\mathrm{AP}} = -a^{*}\mathrm{log}(p_{\mathrm{LLM}}(a|I,H,D)),
\end{equation}
\begin{equation}
\mathcal{L}_{p}=\mathcal{L}_{\mathrm{FI}}+\mathcal{L}_{\mathrm{VIF}}+\mathcal{L}_{\mathrm{AP}},
\end{equation}
where $D$ denotes the textual observations of a single navigation step and $H$ is the history before the step.
$U$, $V$, and $a$ are the output of FI, VIF, AP, respectively.
$U^{*}$, $V^{*}$, and $a^{*}$ represent the ground-truth output extracted from $\mathrm{CoT}_{t}^{*}$ for FI, VIF, and AP, respectively.


\noindent\textbf{Finetuning.}
We reformulate the expert trajectory from the original VLN dataset using  $\mathrm{CoT}_{t}^{*}$ to construct instruction-following finetuning dataset.
At each timestep $t$, 
we train the LLM to generate the complete navigational chain-of-thought $\mathrm{CoT}_{t}$ for action decisions.
The finetuning objective $\mathcal{L}_{f}$ is:
\begin{equation}
\mathcal{L}_{f} = -\sum_{t}\mathrm{CoT}^{*}_{t}\mathrm{log}(p_{\mathrm{LLM}}(\mathrm{CoT_{t}}|I,H_{t},D_{t})).
\end{equation}

After in-domain training, we prompt the trained LLMs to generate the navigational chain-of-thought for action decision based on the prompt and the in-context example described in Sec.~\ref{Design of Chain-of-Thought Prompt}. Benefiting from our introduced in-domain training strategy, the LLM can learn to generate disentangled navigational reasoning with the desired format to guide itself for accurate action decisions.

\begin{table*}[t]

	
	\resizebox{1.0\linewidth}{!}{
	{\renewcommand{\arraystretch}{1.1}
		\begin{tabular}{cc||c|c|c|c|c|c|c|c|c|c}

			\specialrule{.1em}{.05em}{.05em}
				\multirow{2}{*}{Setting}&\multirow{2}{*}{Method}&\multicolumn{5}{c|}{Val Seen}&\multicolumn{5}{c}{Val Unseen}\cr\cline{3-12}
			&&TL&NE $\downarrow$&OSR $\uparrow$&SR $\uparrow$&SPL $\uparrow$&TL&NE $\downarrow$&OSR $\uparrow$&SR $\uparrow$&SPL $\uparrow$\cr
			\hline
			
        
            \multirow{6}{*}{\makecell{cross-modal \\backbone}}&Seq2Seq~\cite{anderson2018vision}&11.33&6.01&53&39&-&8.39&7.81&28&21&-\\
            &Speaker Follower~\cite{fried2018speaker}&-&3.36&74&66&-&-&6.62&45&36&-\\
            &HAMT~\cite{Chen2021HistoryAM}&11.15&2.52&-&76&72&11.46&2.29&-&66&61\\
            &DUET~\cite{Chen2022ThinkGA}&12.32&2.28&86&79&73&13.94&3.31&81&72&60\\
            &BEVBert~\cite{an2022bevbert}&13.56&2.17&88&81&74&14.55&2.81&84&75&64\\
            &ScaleVLN~\cite{wang2023scaling}&13.24&\textbf{2.12}&\textbf{87}&\textbf{81}&\textbf{75}&14.09&\textbf{2.09}&\textbf{88}&\textbf{81}&\textbf{70}\\
            \hline
            \multirow{3}{*}{\makecell{language only \\backbone}}&NavGPT~\cite{zhou2023navgpt}&-&-&-&-&-&11.45&6.46&42&34&29\\
            &NavCoT+LLaMA-Adapter (ours)&-&-&-&-&-&9.19&8.20&28.91&21.97&19.99\\
            &NavCoT+LLaMA 2 (ours)&-&-&-&-&-&9.83&6.67&43.98&36.40&33.17\\
            &NavCoT+LLaMA 2 (ours)*&10.08&\textcolor{blue}{\textbf{6.46}}&\textcolor{blue}{\textbf{48.38}}&\textcolor{blue}{\textbf{41.33}}&\textcolor{blue}{\textbf{38.43}}&9.95&\textcolor{blue}{\textbf{6.26}}&\textcolor{blue}{\textbf{48.11}}&\textcolor{blue}{\textbf{40.23}}&\textcolor{blue}{\textbf{36.64}}\\
            

 \specialrule{.1em}{.05em}{.05em}

		\end{tabular}}}
  \vspace{-0.2cm}
  \caption{Comparison with SOTAs on R2R. * denotes adding randomly chosen 12000 samples from the R2R augmentation dataset~\cite{fried2018speaker}. The best results for cross-modal and language-only backbones are denoted by bold and blue fonts, respectively.}
	\label{tab:com to sota}
	\vspace{-0.2cm}
\end{table*}

	\section{Experiments}

\subsection{Experimental Setup}
\noindent\textbf{Datasets.} 
We evaluate NavCoT on four public VLN benchmarks: R2R~\cite{anderson2018vision}, RxR~\cite{ku2020room}, REVERIE~\cite{qi2020reverie}, and R4R~\cite{jain2019stay}. R2R is built on 90 real-world indoor environments containing 7189 trajectories, each corresponding to three fine-grained instructions. RxR contains much more complex instructions and trajectories than R2R. Since CLIP~\cite{radford2021learning} is pretrained on English language data, we use the English subset of RxR (both en-IN and en-US) for verification, which includes 26464, 2939, 4551 path-instruction pairs for Training, Val Seen, and Val Unseen, respectively. REVERIE replaces the fine-grained instructions in R2R with high-level instructions. R4R concatenates two adjacent tail-to-head trajectories in R2R,
forming longer instructions and trajectories.

\noindent\textbf{Evaluation Metrics.}
The following standard metrics are used for evaluation on R2R~\cite{anderson2018vision} and REVERIE~\cite{qi2020reverie}: 1) Trajectory Length (TL): the average length of the agent's navigated path, 2) Navigation Error (NE): the average distance between the agent's destination and the target viewpoint, 3) Success Rate (SR): the ratio of success, where the agent stops within three meters of the target point, 4) Success rate weighted by Path Length (SPL)~\cite{anderson2018vision}: success rate normalized by the ratio between the length of the shortest path and the predicted path, 5) Oracle Success Rate (OSR): the ratio of containing a viewpoint along the path where the target position is visible. Three evaluation metrics related to the instruction following are added for R4R~\cite{jain2019stay} and RxR~\cite{ku2020room}, i.e., the Coverage weighted
by Length Score (CLS)~\cite{jain2019stay}, the normalized Dynamic Time
Warping (nDTW)~\cite{ilharco2019general}, and the Success weighted by nDTW
(SDTW)~\cite{ilharco2019general}.

\noindent\textbf{Implementation Details.} We train LLaMA-Adapter~\cite{zhang2023llama} and LLaMA 2~\cite{touvron2023llama2} of size 7B with 1.2M and 1.6M trainable parameters, respectively. To speed up the training, we use 4 V100 GPUs with a batch size of 8. The total training time lasts $\sim$10h on 4 V100 GPUs. The inference is conducted on a single V100 GPU. 
We use the AdamW optimizer with the learning rate of 0.001 and the weight decay of 0.02. 
For fast evaluation on various ablation experiments, we randomly choose 90 instruction-trajectory pairs from 8 scans in the val unseen split as the Val Unseen Subset. This subset serves as an efficient testbed to indicate the performance gap among different methods. More implementation details are given in the supplementary material.

\begin{figure*}[t]
\begin{centering}
\includegraphics[width=0.98\linewidth]{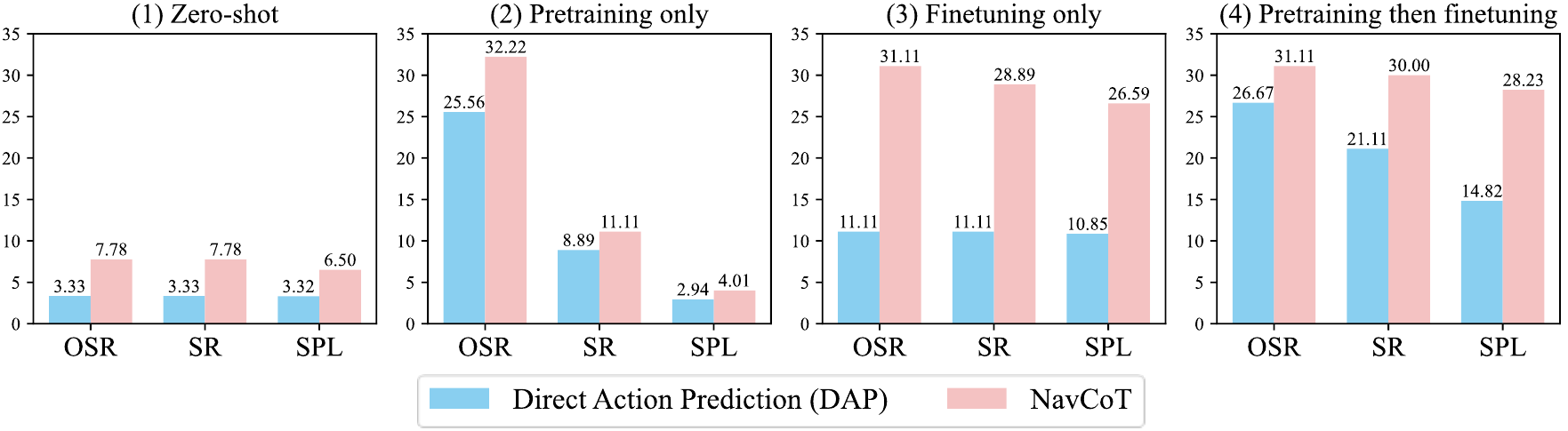}

\par\end{centering}
\vspace{-0.2cm}
\caption{Comparison of NavCoT with the Direct Action Prediction (DAP) variant under different training settings. In DAP, we directly prompt LLM to generate the action prediction. 
}
\vspace{-0.4cm}
\label{fig:ablation}
\end{figure*}

\subsection{Quantitative Results}

\subsubsection{Comparison with Existing Approaches}

Table~\ref{tab:com to sota} presents the comparison between NavCoT and methods with different backbones on R2R~\cite{anderson2018vision}. 
From Table~\ref{tab:com to sota}, we can see that NavCoT with LLaMA-Adapter as backbone shows a large performance gap with NavGPT~\cite{zhou2023navgpt}, revealing the significant
challenge of adapting LLMs with relatively small model
capacity to the VLN task. This shows LLM's original capability is a key factor influencing the navigation performance. However, by plugging into another small language model, i.e., LLaMA2 7B~\cite{touvron2023llama2}, NavCoT outperforms NavGPT which employs GPT-4 as backbone in almost all metrics (4.17\%, 2.4\%, and 1.98\% improvement in SPL,  SR, and OSR, respectively). By
adding a small amount of augmentation data (similar scale
to that of R2R training data), NavCoT further surpasses NavGPT significantly in all metrics (7.64\%, 6.23\%, and 6.11\% improvement in SPL, SR, and OSR, respectively). These results show the effectiveness of our NavCoT in adapting open-source affordable LLMs to the VLN task.

\noindent\textbf{Discussion.}
Note that recent works using LLMs for improving the VLN task, e.g., NavGPT~\cite{zhou2023navgpt}, DiscussNav~\cite{long2023discuss}, and MapGPT~\cite{chen2024mapgpt}, generally show a performance drop compared with previous state-of-the-art VLN methods like scaleVLN~\cite{wang2023scaling}. This demonstrates that the original ability of LLM is far from solving complex navigation tasks. 
Nevertheless, LLM-based agents still have unique advantages for assisting the VLN task.
Firstly, these GPT4-based approaches reveal the potential of zero-shot VLN which is more practical in real-world applications. Moreover,  LLM-based agents can produce navigational reasoning output, which significantly improves decision interpretability and facilitates interaction with humans. These can not be achieved by previous navigation models. 

Besides enhancing the interpretability and interactivity, 
our NavCoT introduces trainable reasoning, 
which adapts a relatively small LLM to the VLN task and shows superiority over GPT4-based methods in improving the accuracy of both navigational reasoning and action decisions. Our proposed method provides a scalable solution to mitigate the domain gap between the LLM's training corpus and the VLN task. 
In the future, it is promising to introduce our NavCoT into large vision-language models (VLMs) to further boost the navigation performance while enhancing interpretability.

\subsubsection{Ablation Study}
In this subsection, we present the ablation results, including training settings, reasoning tasks, chain-of-thought prompts, and backbones, to analyze the effect of each component in NavCoT. More results are given in the supplementary material.


\noindent\textbf{Training settings.}
Fig.~\ref{fig:ablation} presents a comprehensive comparison among various training settings. From Fig.~\ref{fig:ablation}, we can draw three crucial conclusions: 1) NavCoT surpasses the DAP variant in all training settings, highlighting the power of explicit disentangled reasoning; 2) The great superiority of training-based settings over zero-shot ones shows the effectiveness of our parameter-efficient in-domain training strategy; 3) In-domain pretraining and finetuning can encourage the LLM to select actions that are relevant to the target position, reflected in significant improvement of OSR. However, the finetuning contributes much higher to the enhancement of SR and SPL than pretraining by benefiting from the learning of sequential action decisions.

\begin{table}
	
	\centering
			\resizebox{1.0\linewidth}{!}{
	{\renewcommand{\arraystretch}{1.3}\begin{tabular}{c||c|c|c|c|c|c}
			 \specialrule{.1em}{.05em}{.05em}
		
\multirow{2}{*}{Method}&\multicolumn{3}{c|}{pretraining only}&\multicolumn{3}{c}{pretraining \& finetuning}\cr\cline{2-7}
			&OSR $\uparrow$&SR $\uparrow$&SPL $\uparrow$&OSR $\uparrow$&SR $\uparrow$&SPL $\uparrow$\cr
			\hline

          AP&25.56&8.89&2.94&26.67&21.11&14.82\\
          VIF&28.89&10.00&3.43&30.00&21.11&19.08\\
          AP+FI&23.22&8.89&\textbf{4.20}&27.78&24.44&23.78\\
          AP+VIF&\textbf{33.33}&7.78&2.15&24.44&23.33&22.33\\
          
         AP+FI+VIF (ours)&\underline{32.22}&\textbf{11.11}&\underline{4.01}&\textbf{31.11}&\textbf{30.00}&\textbf{28.23}\\
         
          \specialrule{.1em}{.05em}{.05em}
		\end{tabular}}}
\vspace{-0.2cm}	\caption{Ablation results for different proxy tasks on R2R Val Unseen subset.}	
\label{tab:proxy tasks}
	\vspace{-0.2cm}	
\end{table}

\begin{table}
    
\fontsize{14}{14}\selectfont
    \resizebox{1.0\linewidth}{!}{
    {\renewcommand{\arraystretch}{1.2}
    \begin{tabular}{l|cccc}
    \specialrule{.1em}{.05em}{.05em}
			\multirow{2}{*}{Method}&\multicolumn{4}{c}{Val Unseen Subset}\cr\cline{2-5}
   &NE $\downarrow$&OSR $\uparrow$&SR $\uparrow$&SPL $\uparrow$\cr

			\hline

        \multicolumn{4}{l}{$\triangleright$  Chain-of-Thought prompts:}\cr\cline{1-5}
        \hline

          original CoT (GPT-4~\cite{OpenAI_2023})*&7.65&47.78&24.44&12.33\\
         original CoT (LLaMA2)*&9.39&3.33&3.33&1.85\\
        NavCoT (LLaMA2)*&9.02&26.67&16.67&5.79\\
     NavCoT (ours)&\textbf{5.38}&\textbf{58.89}&\textbf{53.33}&\textbf{48.69}\\

 
 
 \hline

        \multicolumn{4}{l}{$\triangleright$   Backbones:}\cr\cline{1-5}
        \hline
        GPT-4*&7.80&22.22&15.56&8.70\\
        LLaMA-Adapter*&9.64&7.78&7.78&6.50\\
        LLaMA-Adapter &7.16&31.11&28.89&26.59\\
        LLaMA 2* &9.02&26.67&16.67&5.79\\
       LLM \& CLIP*&6.95&44.64&32.23&29.14\\
        LLaMA 2 (ours)&\textbf{5.38}&\textbf{58.89}&\textbf{53.33}&\textbf{48.69}\\
        
 \specialrule{.1em}{.05em}{.05em}
    \end{tabular}
}}

\vspace{-0.2cm}
\caption{Ablation results for chain-of-thought prompts and backbones. * denotes the zero-shot manner.}
\label{tab:other ablation}
\vspace{-0.4cm}
\end{table}

\noindent\textbf{Reasoning tasks.}
Table~\ref{tab:proxy tasks} shows how our three reasoning tasks, i.e., Future Imagination (FI), Visual Information Filter (VIF), and Action Prediction (AP) impact on the navigation performance under both pretraining only and pretraining \& finetuning settings. From  Table~\ref{tab:proxy tasks} we can find that the complete combination of three reasoning tasks, i.e., AP+FI+VIF (ours) achieves the comprehensively best performance in both two training settings, showing the effectiveness of our proposed navigational chain of thoughts in improving the action decisions. The results of AP {\it vs.} AP+FI and AP {\it vs.} AP+VIF demonstrate the effects of the reasoning tasks FI and VIF, respectively. Moreover, through the results of AP+FI {\it vs.} AP+FI+VIF (ours) and AP+VIF {\it vs.} AP+FI+VIF (ours), we can also find that both FI and VIF are indispensable in the navigational chain of thoughts.    

\noindent\textbf{Chain-of-thought prompts.}
Table~\ref{tab:other ablation} shows the ablation results for chain-of-thought prompts, 
where we can see that NavCoT outperforms the original CoT, suggesting that our design of combining the world model into CoT can better activate reasonable navigational reasoning. Benefiting from training with formalized labels, NavCoT outperforms both the zero-shot variant and the original CoT for GPT-4~\cite{OpenAI_2023} largely (e.g., $\sim$43 and $\sim$36 points in SPL, respectively). Moreover, note that our NavCoT also shows great superiority than the original CoT for GPT-4~\cite{OpenAI_2023} in the computational efficiency, i.e., the inference time of one-step decision for GPT-4 and NavCoT are $\sim$9.8s and $\sim$0.5s, respectively. 

\noindent\textbf{Backbones.}
We can find in Table~\ref{tab:other ablation} that NavCoT+LLaMA-Adapter is largely inferior to its alternative NavCoT+LLaMA2. This is not surprising since LLaMA-Adapter is based on LLaMA 1~\cite{touvron2023llama} while LLaMA 2 is trained on 40\% more data than the previous version. Moreover, LLaMA-Adapter has 0.4M fewer trainable parameters than the bias tuning setting~\cite{gao2023llama}. However, NavCoT shows consistent improvement on both backbones under our in-domain training strategies. Interestingly, we can find that LLaMA 2 outperforms GPT-4  in the zero-shot NavCoT setting, probably because when facing the prompt with less information and constrained output format, the advantage of GPT-4 is not obvious over the smaller language model LLaMA 2. However, such kind of prompt is more beneficial for implementing in-domain training. 

We also present the result of using the imagination and CLIP-similarity to make action decisions, which we name as  ``LLM \& CLIP*''. Concretely, we utilize the imagination, i.e., imagined landmark generated by NavCoT (LLaMA2) to select the observation (action) by calculating the CLIP-smilarity between the textual landmark feature and the observation feature. From Table~\ref{tab:other ablation}, we can find that LLM \& CLIP* outperforms LLaMA 2*, LLaMA-Adapter, LLaMA-Adapter*, and GPT-4*, showing that using CLIP for matching the imagination (i.e., the textually represented landmark) with the right observations has relatively high accuracy for the action decision. This demonstrates the reasonability of our introduced strategy for collecting imagination ground-truths (GTs) using LLM and CLIP. We can also find that LLaMA 2 (ours) surpasses LLM \& CLIP* by a large margin. This is reasonable since the navigation decision relies on comprehensive factors including landmarks, navigation history, direction, {\it etc}. This result further reveals the importance of our parameter-efficient in-domain training strategy that effectively adapts the LLM backbone for making complex navigation decisions. 

\begin{table}
    
\fontsize{17}{17}\selectfont
    \resizebox{1.0\linewidth}{!}{
    {\renewcommand{\arraystretch}{1.2}
    \begin{tabular}{l|ccccc}
    \specialrule{.1em}{.05em}{.05em}
			\multirow{2}{*}{Method}&\multicolumn{5}{c}{Val Unseen}\cr\cline{2-6}
&SR $\uparrow$&SPL $\uparrow$&CLS $\uparrow$&nDTW $\uparrow$&SDTW $\uparrow$\cr
			\hline
			
        EnvDrop~\cite{tan2019learning}&38.5&34&54&51&32\\
        HAMT*~\cite{Chen2021HistoryAM}&\textbf{38.26}&\textbf{36.23}&\textbf{58.45}&\textbf{53.08}&\textbf{32.81}\\
        \hline
        Direct Action Prediction (DAP)&20.46&18.68&37.19&33.64&16.02\\
        NavCoT (ours)&\textcolor{blue}{\textbf{24.52}}&\textcolor{blue}{\textbf{22.58}}&\textcolor{blue}{\textbf{45.06}}&\textcolor{blue}{\textbf{38.94}}&\textcolor{blue}{\textbf{19.63}}\\

 \specialrule{.1em}{.05em}{.05em}
    \end{tabular}
}}

\vspace{-0.2cm}
\caption{Comparison results on RxR English subset. * denotes our re-implementation results with the imitation learning strategy~\cite{Chen2021HistoryAM}. The best results for cross-modal and language-only backbones are denoted by bold and blue fonts, respectively.}
\vspace{-0.2cm}
\label{tab:rxr}
\end{table}

\begin{table}
    
\fontsize{17}{17}\selectfont
    \resizebox{1.0\linewidth}{!}{
    {\renewcommand{\arraystretch}{1.2}
    \begin{tabular}{l|cccc}
    \specialrule{.1em}{.05em}{.05em}
			\multirow{2}{*}{Method}&\multicolumn{4}{c}{Val Unseen}\cr\cline{2-5}
&TL&SR $\uparrow$&OSR $\uparrow$&SPL $\uparrow$\cr			\hline
			

        Seq2Seq~\cite{qi2020reverie}&-&4.2&9.07&2.84\\
        RCM~\cite{wang2019reinforced}&11.98&9.29&14.23&6.97\\
        FAST-MATTN~\cite{qi2020reverie}&45.28&14.40&28.20&7.19\\
        HAMT*~\cite{Chen2021HistoryAM}&9.24&\textbf{23.80}&\textbf{26.44}&\textbf{22.42}\\
        \hline
        Direct Action Prediction (DAP)&16.30&3.12&7.36&1.74\\
        NavCoT (ours)&12.36&\textcolor{blue}{\textbf{9.20}}&\textcolor{blue}{\textbf{14.20}}&\textcolor{blue}{\textbf{7.18}}\\

 \specialrule{.1em}{.05em}{.05em}
    \end{tabular}
}}

\vspace{-0.2cm}
\caption{Comparison results on REVERIE. * denotes our re-implementation results with the imitation learning strategy~\cite{Chen2021HistoryAM}. The best results for cross-modal and language-only backbones are denoted by bold and blue fonts, respectively.}
\vspace{-0.4cm}
\label{tab:reverie}
\end{table}

\subsubsection{Generalization on Other Datasets}

To further verify the generalization of NavCoT on different VLN datasets with much longer instructions (RxR~\cite{ku2020room}) and high-level instructions (REVERIE~\cite{qi2020reverie}), we conduct experiments on RxR and REVERIE, and the results are given in Table~\ref{tab:rxr} and Table~\ref{tab:reverie}, respectively. Since there are no reported results of language-only backbone on RxR and REVERIE, we develop a naive baseline, which is a variant of NavCoT that makes direct action predictions (DAP) rather than generating navigational chain-of-thoughts.
We also present the results of a strong pretrained cross-modal backbone HAMT~\cite{Chen2021HistoryAM} with the same training strategy as ours. 
From Table~\ref{tab:rxr} we can find that although NavCoT shows the performance gap with the cross-modal methods, it significantly outperforms DAP especially in the CLS and nDTW metrics, showing that our method not only improves the navigation accuracy but also enables better instruction following, which is crucial for long-horizon navigation. Table~\ref{tab:reverie} shows that different methods encounter significant performance drop on REVERIE compared to R2R, revealing the challenge of navigating under high-level instructions with limited information. However, NavCoT is still superior over DAP in a large margin, showing the effectiveness of the proposed method. 

\begin{table*}[t]
    
    \centering
    \resizebox{0.8\linewidth}{!}{
    {\renewcommand{\arraystretch}{1.2}
    \begin{tabular}{l|cccc|cccc}
    \specialrule{.1em}{.05em}{.05em}
			\multirow{2}{*}{Method}&\multicolumn{4}{c}{R4R}&\multicolumn{4}{c}{R2R}\cr\cline{2-9}
   &OSR $\uparrow$&SR $\uparrow$&CLS$\uparrow$&NDTW$\uparrow$&NE$\downarrow$&OSR $\uparrow$&SR $\uparrow$&SPL $\uparrow$\cr

			\hline
			
        
        HAMT*~\cite{Chen2021HistoryAM}&53.80&22.20&54.29&39.49&4.74&63.47&57.43&54.8\\
        \hline
        Direct Action Prediction (DAP)&25.80&12.80&19.73&16.88&7.89&23.50&20.52&19.96\\
        NavCoT (ours)&42.20&13.00&45.99&31.79&7.08&37.46&31.33&29.08\\

 \specialrule{.1em}{.05em}{.05em}
    \end{tabular}
}}

\vspace{-0.2cm}
\caption{Low-resource experimental results on Val Unseen on R4R and R2R. * denotes our re-implementation results with the imitation learning strategy~\cite{Chen2021HistoryAM}. In R4R, CLS and NDTW are the main metrics.}
\vspace{-0.4cm}
\label{tab:r4r}
\end{table*}

\begin{figure*}[t]
\begin{centering}
\includegraphics[width=0.95\linewidth]{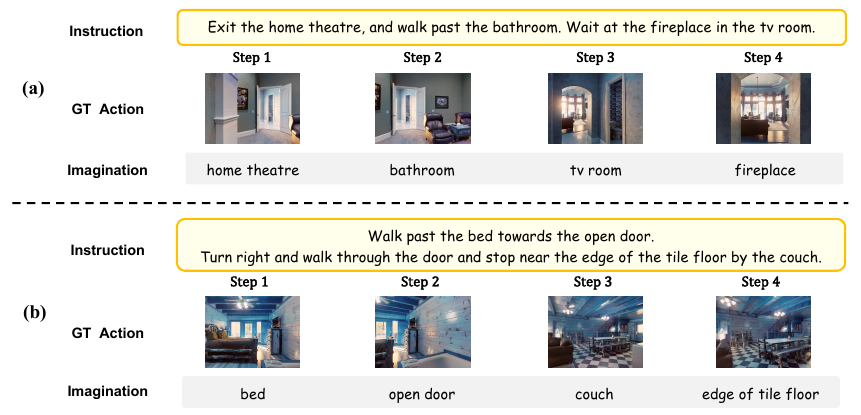}
\par\end{centering}
\vspace{-0.2cm}
\caption{Visualization examples of Imagination ground-truth (GT). We do not show the imagination GT for the final step which is ``stop''.
}\label{fig:visualiation_imagination_supp}
\vspace{-0.2cm}
\end{figure*}

\begin{figure*}[h]
\begin{centering}
\includegraphics[width=0.98\linewidth]{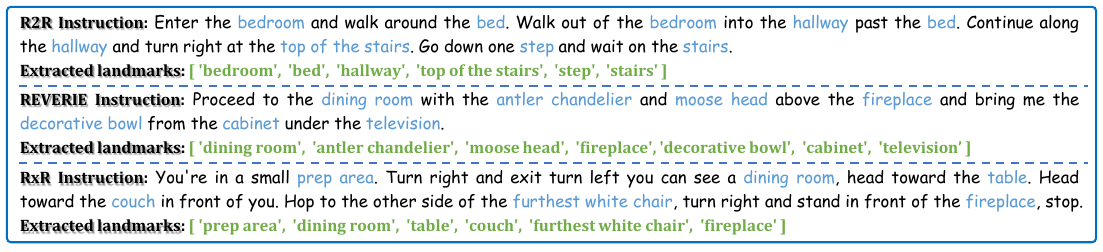}

\par\end{centering}
\vspace{-0.2cm}
\caption{Landmark extraction visualization of different datasets.}
\vspace{-0.4cm}
\label{fig:landmark_accuracy}
\end{figure*}

\begin{figure*}[t]
\begin{centering}
\includegraphics[width=0.95\linewidth]{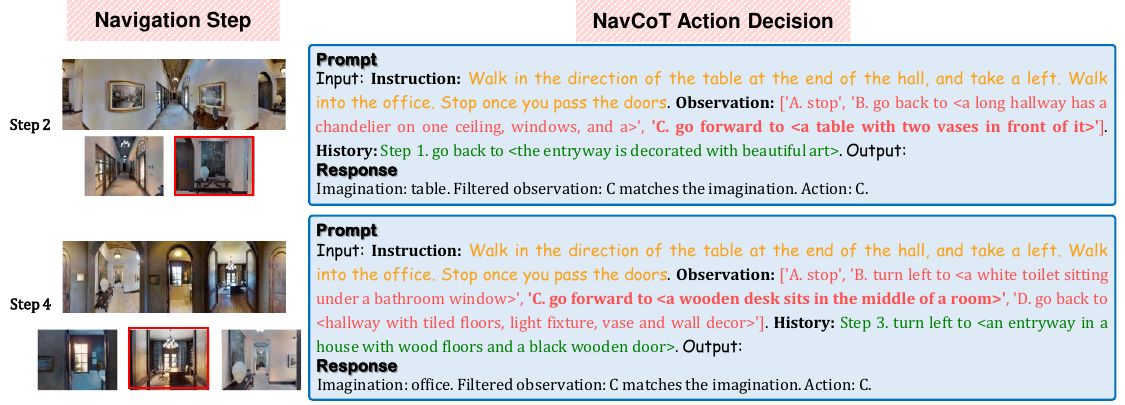}
\par\end{centering}
\vspace{-0.2cm}
\caption{Action decision visualization of NavCoT. We only extract two steps for illustration and give the whole trajectory in the supplementary material. The ground-truth actions are annotated by red boxes.
}
\vspace{-0.2cm}
\label{fig:visualiation_action}
\end{figure*}

\subsubsection{Low-Resource Experiments}

We further conduct the low-resource experiments on two popular VLN benchmarks R2R~\cite{anderson2018vision} and R4R~\cite{jain2019stay} to validate the ability of NavCoT when facing a small amount of training data. R4R contains much longer instructions and trajectories than R2R. For R4R, 
we randomly choose 5000 trajectories from 50 scans in Train for training and 500 trajectories from 11 scans in Val Unseen for validation. For R2R, we randomly extract 3000 trajectories from 61 scans in Train for training. We compare NavCoT with Direct Action Prediction (DAP)  and a strong cross-modal backbone method HAMT~\cite{Chen2021HistoryAM}. 

Table~\ref{tab:r4r} shows that under the low-resource setting with the same training strategy, the performance gap between NavCoT and HAMT~\cite{Chen2021HistoryAM} is largely reduced compared to that in Table~\ref{tab:com to sota}. For example, the performance gap of CLS between HAMT~\cite{Chen2021HistoryAM} and NavCoT is reduced to $\sim$8.  Moreover, NavCoT outperforms DAP in a large margin on both R2R and R4R, showing the good generalization ability of NavCoT under low-resource settings. Note that the results on R4R show that NavCoT surpasses DAP significantly in OSR, CLS, and NDTW, demonstrating that beyond correct action decisions, NavCoT also shows promising instruction following ability for long trajectories.

\begin{figure*}[t]
\begin{centering}
\includegraphics[width=0.98\linewidth]{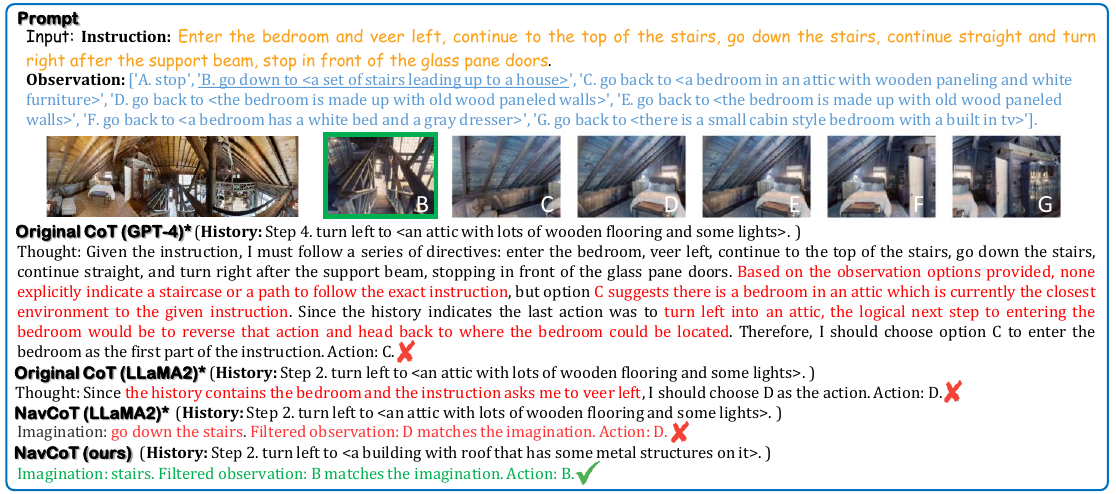}

\par\end{centering}
\vspace{-0.2cm}
\caption{Reasoning output visualization of different Chain-of-Thought (CoT) methods. Apart from NavCoT (ours), all other comparison methods are zero-shot.}
\vspace{-0.4cm}
\label{fig:comparison_cot}
\end{figure*}

\begin{figure}
\begin{centering}
\includegraphics[width=1.0\linewidth]{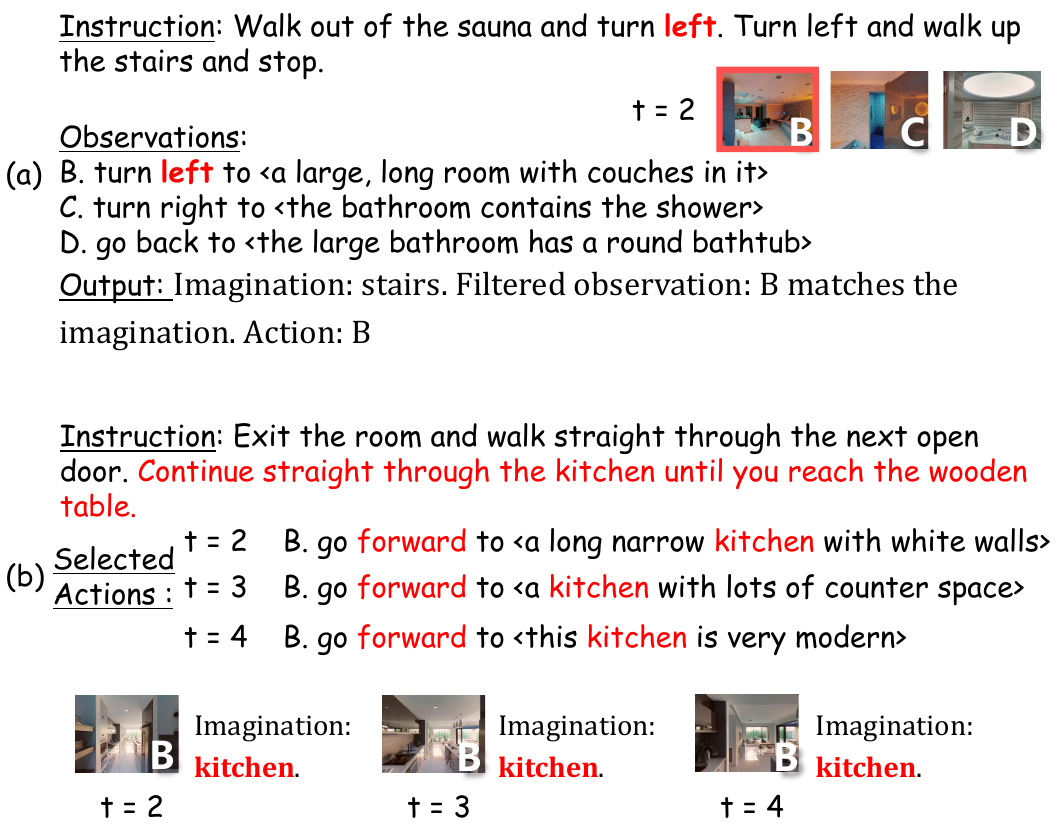}
\par\end{centering}
\vspace{-0.2cm}
\caption{Generalization to special cases: (a) Little landmarks: There are little landmarks mentioned in the instruction. (b) Same landmark: multiple continuous navigation steps point to the same landmark. 
}
\label{fig:prompt}
\vspace{-0.2cm}
\end{figure}

\subsection{Visualization}
\noindent\textbf{Quality of Imagination ground-truths (GTs).}
Fig.~\ref{fig:visualiation_imagination_supp} gives some visualization examples of imagination GT. In Fig.~\ref{fig:visualiation_imagination_supp}, we can observe that the application of LLM and CLIP in the ground-truth collection process effectively ensures the quality of the imagination GT even if when the landmark is rarely mentioned in the corpus (Step 4 in Fig.~\ref{fig:visualiation_imagination_supp}(b)) and the landmark only occupies a small region in the observation (Step 2 in Fig.~\ref{fig:visualiation_imagination_supp}(b) and Step 4 in Fig.~\ref{fig:visualiation_imagination_supp}(a)). We also conduct a quantitative human evaluation by randomly extracting 200 trajectories on R2R. The accuracy of CLIP reaches $\sim$ 73\%, which is a relatively high value. Note that since the decisions are made based on comprehensive information, a small proportion of noisy imagination labels are tolerable.

Since we utilize the LLM for instruction landmark extraction to collect imagination GTs, we further validate the accuracy of landmark extraction. Specifically, we manually check the landmark extraction results of our used LLM, i.e., 
ChatGPT~\cite{openai} for the R2R, REVERIE and RxR datasets. For each dataset, we randomly select 100 instructions for evaluation. The landmark extraction is considered successful if every landmark in it is extracted. We find that almost all landmarks can be successfully extracted by the ChatGPT, and the success rate of landmark extraction is 100\% for R2R, 98\% for REVERIE, and 98\% for RxR. Since the landmarks of an R4R instruction can be combined from the landmarks of the corresponding R2R instructions, the landmark extraction accuracy of R4R instructions is consistent with that of R2R instructions. We also present the visualization results of landmark extraction for different datasets in Fig.~\ref{fig:landmark_accuracy}, where we can observe that ChatGPT extracts landmarks accurately for instructions of different datasets, demonstrating that it is robust to instructions with different styles. The accurate landmark extraction effectively ensures the correctness of our subsequent collection of the Imagination GTs.

\noindent\textbf{Action decision visualizations.}
Fig.~\ref{fig:visualiation_action} gives the action decision visualization of NavCoT, where we can find that NavCoT generates reasonable navigational reasoning for guiding itself to make action decisions. For instance, in Step 2, from the history and observations, NavCoT infers that its position is possibly the \textit{hall} and the option C may contain the desired \textit{table}, then outputs the imagination that leads to the correct action. Notably, due to the domain gap between NavCoT and VLM, the observations inevitably do not always contain the imagination. However, with in-domain training, NavCoT shows the emerging ability to recognize potential connection between observations and imaginations. For example, in Step 4, the imagination \textit{office} does not directly match any observation,  whereas NavCoT correctly chooses the option C containing a \textit{wooden desk}  which suggests a scene of office.

\noindent\textbf{Comparison of different CoT methods.}
Figure~\ref{fig:comparison_cot} presents the visualization comparison of our NavCoT with other three zero-shot CoT methods: the original CoT (GPT-4)*, the original CoT (LLaMA2)*, and the NavCoT (LLaMA2)*. From Figure~\ref{fig:comparison_cot}, we can observe two obvious advantages of our NavCoT over the other CoT approaches: 1) Through formalizing navigational reasoning into three key steps: future imagination, visual information filtering, and action decision, NavCoT significantly reduces the redundant information caused by the domain gap between the LLM training corpus and VLN task in its reasoning output.  In contrast, the reasoning of the original CoT (GPT-4)* contains extensive redundancy, and the reasoning of the original CoT (LLaMA2)* contains a large amount of noise. For instance, the original CoT (GPT-4)* mixes the instruction, common sense, and history in its output, and produces redundant reasoning which is hard to extract useful information from it. The original CoT (LLaMA2)* outputs the wrong reasoning that ``the history contains the bedroom'' and its chosen action D (the direction is go back) does not match its reasoning that ``the instruction asks me to veer left''. 2) By introducing the parameter-efficient in-domain training strategies, NavCoT can produce reasonable formalized imagination and uses the imagination to guide itself for choosing the correct action decision. For example, our NavCoT produces proper imagination ``stairs'' and matches the imagination to  the correct option B which contains the stairs for successful navigation. In contrast, the NavCoT (LLaMA2)* fails to match its imagination output ``go down the stairs'' to the correct option B and instead chooses option D.

\noindent\textbf{Generalization to special cases.} We present some visualization examples in Fig.~\ref{fig:prompt} to show how NavCoT generalizes to typical special cases in VLN, e.g., there are little landmarks mentioned in the instruction or multiple continuous navigation steps pointing to the same landmark. From Fig.~\ref{fig:prompt}, we can find that NavCoT is capable of making correct navigation decisions under these cases. For example, in Fig.~\ref{fig:prompt}(a), even if the instruction contain little landmarks and limit the effect of the imagination, through the direction information provided by our vision-to-text system, NavCoT can still perform direction-level cross-modal alignment between the observation description and the instruction. Moreover, since the training data naturally contain many cases where continuous steps point to the same landmark, NavCoT can learn to generate reasonable imagination after training, as shown in Fig.~\ref{fig:prompt}(b).

	\section{Conclusion}
This work  introduces NavCoT, which fulfills parameter-efficient in-domain training for enabling LLMs to perform self-guided navigational decisions.
Experimental results show the great superiority of NavCoT over a recent high-cost LLM-based VLN approach and direct action prediction variants.
We believe that our method makes a solid step towards developing scalable LLM-based VLN approaches and provides a meaningful reference in designing trainable navigational reasoning generation strategies for improving both the accuracy and interpretability of action decision.
Constrained by the detail information loss during the vision-to-text transformation, the LLM may fail to make accurate decisions in some cases.  Future direction includes introducing our NavCoT into powerful large vision-language models to further improve the navigation performance.
	
	
	%


	\section*{Acknowledgment}
	

        This work was supported by National Key Research and Development Program of China (2024YFE0203100), National Natural Science Foundation of China (NSFC) under Grants No.62476293, Shenzhen Science and Technology Program (Grant No.GJHZ20220913142600001), Nansha Key R\&D Program under Grant No.2022ZD014, and General Embodied AI Center of Sun Yat-sen University.
	
	\ifCLASSOPTIONcaptionsoff
	\newpage
	\fi

	
	
	%
		
		
\bibliographystyle{IEEEtran}
\bibliography{IEEEabrv,egbib}	

\begin{thebibliography}{10}
\providecommand{\url}[1]{#1}
\csname url@samestyle\endcsname
\providecommand{\newblock}{\relax}
\providecommand{\bibinfo}[2]{#2}
\providecommand{\BIBentrySTDinterwordspacing}{\spaceskip=0pt\relax}
\providecommand{\BIBentryALTinterwordstretchfactor}{4}
\providecommand{\BIBentryALTinterwordspacing}{\spaceskip=\fontdimen2\font plus
\BIBentryALTinterwordstretchfactor\fontdimen3\font minus \fontdimen4\font\relax}
\providecommand{\BIBforeignlanguage}[2]{{%
\expandafter\ifx\csname l@#1\endcsname\relax
\typeout{** WARNING: IEEEtran.bst: No hyphenation pattern has been}%
\typeout{** loaded for the language `#1'. Using the pattern for}%
\typeout{** the default language instead.}%
\else
\language=\csname l@#1\endcsname
\fi
#2}}
\providecommand{\BIBdecl}{\relax}
\BIBdecl

\bibitem{anderson2018vision}
P.~{Anderson}, Q.~{Wu}, D.~{Teney}, J.~{Bruce}, M.~{Johnson}, N.~{Sunderhauf}, I.~{Reid}, S.~{Gould}, and A.~van~den {Hengel}, ``Vision-and-language navigation: Interpreting visually-grounded navigation instructions in real environments,'' in \emph{CVPR}, 2018.

\bibitem{qi2020reverie}
Y.~{Qi}, Q.~{Wu}, P.~{Anderson}, X.~{Wang}, W.~Y. {Wang}, C.~{Shen}, and A.~van~den {Hengel}, ``Reverie: Remote embodied visual referring expression in real indoor environments,'' in \emph{CVPR}, 2020.

\bibitem{Chen2019TOUCHDOWNNL}
H.~Chen, A.~Suhr, D.~K. Misra, N.~Snavely, and Y.~Artzi, ``Touchdown: Natural language navigation and spatial reasoning in visual street environments,'' in \emph{CVPR}, 2019.

\bibitem{jain2019stay}
V.~{Jain}, G.~{Magalhaes}, A.~{Ku}, A.~{Vaswani}, E.~{Ie}, and J.~{Baldridge}, ``Stay on the path: Instruction fidelity in vision-and-language navigation,'' in \emph{ACL}, 2019.

\bibitem{ku2020room}
A.~{Ku}, P.~{Anderson}, R.~{Patel}, E.~{Ie}, and J.~{Baldridge}, ``Room-across-room: Multilingual vision-and-language navigation with dense spatiotemporal grounding,'' in \emph{EMNLP}, 2020.

\bibitem{brown2020language}
T.~Brown, B.~Mann, N.~Ryder, M.~Subbiah, J.~D. Kaplan, P.~Dhariwal, A.~Neelakantan, P.~Shyam, G.~Sastry, A.~Askell, and et~al., ``Language models are few-shot learners,'' in \emph{NeurIPS}, 2020.

\bibitem{touvron2023llama}
H.~Touvron, T.~Lavril, G.~Izacard, X.~Martinet, M.-A. Lachaux, T.~Lacroix, B.~Rozi{\`e}re, N.~Goyal, E.~Hambro, F.~Azhar \emph{et~al.}, ``Llama: Open and efficient foundation language models,'' \emph{arXiv preprint arXiv:2302.13971}, 2023.

\bibitem{touvron2023llama2}
H.~Touvron, L.~Martin, K.~Stone, P.~Albert, A.~Almahairi, Y.~Babaei, N.~Bashlykov, S.~Batra, P.~Bhargava, S.~Bhosale \emph{et~al.}, ``Llama 2: Open foundation and fine-tuned chat models,'' \emph{arXiv preprint arXiv:2307.09288}, 2023.

\bibitem{Ahn2022DoAI}
M.~Ahn, A.~Brohan, N.~Brown, Y.~Chebotar, O.~Cortes, B.~David, C.~Finn, C.~Fu, K.~Gopalakrishnan \emph{et~al.}, ``Do as i can and not as i say: Grounding language in robotic affordances,'' \emph{arXiv preprint arXiv:2204.01691}, 2022.

\bibitem{Huang2022InnerME}
W.~Huang, F.~Xia, T.~Xiao, H.~Chan, J.~Liang, P.~R. Florence, A.~Zeng, J.~Tompson, I.~Mordatch, Y.~Chebotar, P.~Sermanet, N.~Brown, T.~Jackson, L.~Luu, S.~Levine, K.~Hausman, and B.~Ichter, ``Inner monologue: Embodied reasoning through planning with language models,'' \emph{arXiv preprint arXiv:2207.05608}, 2022.

\bibitem{driess2023palme}
D.~Driess, F.~Xia, M.~S.~M. Sajjadi, C.~Lynch, A.~Chowdhery, B.~Ichter, A.~Wahid, J.~Tompson, Q.~Vuong, T.~Yu \emph{et~al.}, ``Palm-e: An embodied multimodal language model,'' \emph{arXiv preprint arXiv:2303.03378}, 2023.

\bibitem{li2023blip2}
J.~Li, D.~Li, S.~Savarese, and S.~Hoi, ``{BLIP-2:} bootstrapping language-image pre-training with frozen image encoders and large language models,'' in \emph{ICML}, 2023.

\bibitem{li2022blip}
J.~Li, D.~Li, C.~Xiong, and S.~Hoi, ``Blip: Bootstrapping language-image pre-training for unified vision-language understanding and generation,'' in \emph{ICML}, 2022.

\bibitem{zhou2023navgpt}
G.~Zhou, Y.~Hong, and Q.~Wu, ``Navgpt: Explicit reasoning in vision-and-language navigation with large language models,'' in \emph{AAAI}, 2024.

\bibitem{long2023discuss}
Y.~Long, X.~Li, W.~Cai, and H.~Dong, ``Discuss before moving: Visual language navigation via multi-expert discussions,'' \emph{arXiv preprint arXiv:2309.11382}, 2023.

\bibitem{OpenAI_2023}
O.~OpenAI, ``\BIBforeignlanguage{en-US}{Gpt-4 technical report},'' Mar 2023.

\bibitem{johnson1983mental}
P.~N. Johnson-Laird, \emph{Mental models: Towards a cognitive science of language, inference, and consciousness}.\hskip 1em plus 0.5em minus 0.4em\relax Harvard University Press, 1983, no.~6.

\bibitem{johnson2010mental}
{Johnson-Laird, Philip N}, ``Mental models and human reasoning,'' \emph{Proceedings of the National Academy of Sciences}, vol. 107, no.~43, pp. 18\,243--18\,250, 2010.

\bibitem{wei2022chain}
J.~Wei, X.~Wang, D.~Schuurmans, M.~Bosma, F.~Xia, E.~Chi, Q.~V. Le, D.~Zhou \emph{et~al.}, ``Chain-of-thought prompting elicits reasoning in large language models,'' in \emph{NeurIPS}, 2022.

\bibitem{zhang2023llama}
R.~Zhang, J.~Han, A.~Zhou, X.~Hu, S.~Yan, P.~Lu, H.~Li, P.~Gao, and Y.~Qiao, ``Llama-adapter: Efficient fine-tuning of language models with zero-init attention,'' \emph{arXiv preprint arXiv:2303.16199}, 2023.

\bibitem{tan2019learning}
H.~{Tan}, L.~{Yu}, and M.~{Bansal}, ``Learning to navigate unseen environments: Back translation with environmental dropout,'' in \emph{NAACL-HLT}, 2019.

\bibitem{fried2018speaker}
D.~{Fried}, R.~{Hu}, V.~{Cirik}, A.~{Rohrbach}, J.~{Andreas}, L.-P. {Morency}, T.~{Berg-Kirkpatrick}, K.~{Saenko}, D.~{Klein}, and T.~{Darrell}, ``Speaker-follower models for vision-and-language navigation,'' in \emph{NeurIPS}, 2018.

\bibitem{liu2021vision}
C.~{Liu}, F.~{Zhu}, X.~{Chang}, X.~{Liang}, Z.~{Ge}, and Y.-D. {Shen}, ``Vision-language navigation with random environmental mixup,'' in \emph{ICCV}, 2021.

\bibitem{Fu2019CounterfactualVN}
T.-J. Fu, X.~E. Wang, M.~F. Peterson, S.~T. Grafton, M.~P. Eckstein, and W.~Y. Wang, ``Counterfactual vision-and-language navigation via adversarial path sampling,'' in \emph{ECCV}, 2020.

\bibitem{liang2022contrastive}
X.~Liang, F.~Zhu, Y.~Zhu, B.~Lin, B.~Wang, and X.~Liang, ``Contrastive instruction-trajectory learning for vision-language navigation,'' in \emph{Proceedings of the AAAI Conference on Artificial Intelligence}, vol.~36, no.~2, 2022, pp. 1592--1600.

\bibitem{zhu2020vision}
F.~{Zhu}, Y.~{Zhu}, X.~{Chang}, and X.~{Liang}, ``Vision-language navigation with self-supervised auxiliary reasoning tasks,'' in \emph{CVPR}, 2020.

\bibitem{lin2021adversarial}
B.~Lin, Y.~Zhu, Y.~Long, X.~Liang, Q.~Ye, and L.~Lin, ``Adversarial reinforced instruction attacker for robust vision-language navigation,'' \emph{IEEE Transactions on Pattern Analysis and Machine Intelligence}, vol.~44, no.~10, pp. 7175--7189, 2021.

\bibitem{lin2022adapt}
B.~Lin, Y.~Zhu, Z.~Chen, X.~Liang, J.~Liu, and X.~Liang, ``Adapt: Vision-language navigation with modality-aligned action prompts,'' in \emph{Proceedings of the IEEE/CVF Conference on Computer Vision and Pattern Recognition}, 2022, pp. 15\,396--15\,406.

\bibitem{wang2019reinforced}
X.~{Wang}, Q.~{Huang}, A.~{Celikyilmaz}, J.~{Gao}, D.~{Shen}, Y.-F. {Wang}, W.~Y. {Wang}, and L.~{Zhang}, ``Reinforced cross-modal matching and self-supervised imitation learning for vision-language navigation,'' in \emph{CVPR}, 2019.

\bibitem{ma2019self}
C.-Y. {Ma}, jiasen {lu}, Z.~{Wu}, G.~{AlRegib}, Z.~{Kira}, richard {socher}, and C.~{Xiong}, ``Self-monitoring navigation agent via auxiliary progress estimation,'' in \emph{ICLR}, 2019.

\bibitem{deng2020evolving}
Z.~{Deng}, K.~{Narasimhan}, and O.~{Russakovsky}, ``Evolving graphical planner: Contextual global planning for vision-and-language navigation,'' in \emph{NeurIPS}, 2020.

\bibitem{qi2020Object}
Y.~{Qi}, Z.~{Pan}, S.~{Zhang}, A.~van~den {Hengel}, and Q.~{Wu}, ``Object-and-action aware model for visual language navigation,'' in \emph{ECCV}, 2020.

\bibitem{hong2021vln}
Y.~{Hong}, Q.~{Wu}, Y.~{Qi}, C.~{Rodriguez-Opazo}, and S.~{Gould}, ``Vln bert: A recurrent vision-and-language bert for navigation,'' in \emph{CVPR}, 2021.

\bibitem{Chen2021HistoryAM}
S.~Chen, P.-L. Guhur, C.~Schmid, and I.~Laptev, ``History aware multimodal transformer for vision-and-language navigation,'' in \emph{NeurIPS}, 2021.

\bibitem{Chen2022ThinkGA}
S.~Chen, P.-L. Guhur, M.~Tapaswi, C.~Schmid, and I.~Laptev, ``Think global, act local: Dual-scale graph transformer for vision-and-language navigation,'' in \emph{CVPR}, 2022.

\bibitem{Qiao2022HOPHA}
Y.~Qiao, Y.~Qi, Y.~Hong, Z.~Yu, P.~Wang, and Q.~Wu, ``Hop: History-and-order aware pre-training for vision-and-language navigation,'' in \emph{CVPR}, 2022.

\bibitem{Guhur2021AirbertIP}
P.-L. Guhur, M.~Tapaswi, S.~Chen, I.~Laptev, and C.~Schmid, ``Airbert: In-domain pretraining for vision-and-language navigation,'' in \emph{ICCV}, 2021.

\bibitem{an2022bevbert}
D.~An, Y.~Qi, Y.~Li, Y.~Huang, L.~Wang, T.~Tan, and J.~Shao, ``Bevbert: Topo-metric map pre-training for language-guided navigation,'' in \emph{ICCV}, 2023.

\bibitem{wang2023scaling}
Z.~Wang, J.~Li, Y.~Hong, Y.~Wang, Q.~Wu, M.~Bansal, S.~Gould, H.~Tan, and Y.~Qiao, ``Scaling data generation in vision-and-language navigation,'' in \emph{ICCV}, 2023.

\bibitem{chen2024mapgpt}
J.~Chen, B.~Lin, R.~Xu, Z.~Chai, X.~Liang, and K.-Y. Wong, ``Mapgpt: Map-guided prompting with adaptive path planning for vision-and-language navigation,'' in \emph{Proceedings of the 62nd Annual Meeting of the Association for Computational Linguistics (Volume 1: Long Papers)}, 2024, pp. 9796--9810.

\bibitem{chen2024affordances}
J.~Chen, B.~Lin, X.~Liu, L.~Ma, X.~Liang, and K.-Y.~K. Wong, ``Affordances-oriented planning using foundation models for continuous vision-language navigation,'' \emph{arXiv preprint arXiv:2407.05890}, 2024.

\bibitem{zheng2024towards}
D.~Zheng, S.~Huang, L.~Zhao, Y.~Zhong, and L.~Wang, ``Towards learning a generalist model for embodied navigation,'' in \emph{Proceedings of the IEEE/CVF Conference on Computer Vision and Pattern Recognition}, 2024, pp. 13\,624--13\,634.

\bibitem{Yao2022ReActSR}
S.~Yao, J.~Zhao, D.~Yu, N.~Du, I.~Shafran, K.~Narasimhan, and Y.~Cao, ``React: Synergizing reasoning and acting in language models,'' in \emph{ICLR}, 2023.

\bibitem{shahlm}
D.~Shah, B.~Osinski, B.~Ichter, and S.~Levine, ``Lm-nav: Robotic navigation with large pre-trained models of language, vision, and action,'' in \emph{CoRL}, 2022.

\bibitem{schumann-2023-velma}
R.~Schumann, W.~Zhu, W.~Feng, T.-J. Fu, S.~Riezler, and W.~Y. Wang, ``Velma: Verbalization embodiment of llm agents for vision and language navigation in street view,'' \emph{arXiv preprint arXiv:2307.06082}, 2023.

\bibitem{wang2023voyager}
G.~Wang, Y.~Xie, Y.~Jiang, A.~Mandlekar, C.~Xiao, Y.~Zhu, L.~Fan, and A.~Anandkumar, ``Voyager: An open-ended embodied agent with large language models,'' \emph{arXiv preprint arXiv:2305.16291}, 2023.

\bibitem{mu2023embodiedgpt}
Y.~Mu, Q.~Zhang, M.~Hu, W.~Wang, M.~Ding, J.~Jin, B.~Wang, J.~Dai, Y.~Qiao, and P.~Luo, ``Embodiedgpt: Vision-language pre-training via embodied chain of thought,'' \emph{arXiv preprint arXiv:2305.15021}, 2023.

\bibitem{yang2023octopus}
J.~Yang, Y.~Dong, S.~Liu, B.~Li, Z.~Wang, C.~Jiang, H.~Tan, J.~Kang, Y.~Zhang, K.~Zhou \emph{et~al.}, ``Octopus: Embodied vision-language programmer from environmental feedback,'' \emph{arXiv preprint arXiv:2310.08588}, 2023.

\bibitem{brohan2023rt}
B.~Zitkovich, T.~Yu, S.~Xu, P.~Xu, T.~Xiao, F.~Xia, J.~Wu, P.~Wohlhart, S.~Welker, A.~Wahid \emph{et~al.}, ``Rt-2: Vision-language-action models transfer web knowledge to robotic control,'' in \emph{CoRL}, 2023.

\bibitem{wang2022self}
X.~Wang, J.~Wei, D.~Schuurmans, Q.~Le, E.~Chi, S.~Narang, A.~Chowdhery, and D.~Zhou, ``Self-consistency improves chain of thought reasoning in language models,'' in \emph{ICLR}, 2023.

\bibitem{zhou2022least}
D.~Zhou, N.~Sch{\"a}rli, L.~Hou, J.~Wei, N.~Scales, X.~Wang, D.~Schuurmans, C.~Cui, O.~Bousquet, Q.~Le \emph{et~al.}, ``Least-to-most prompting enables complex reasoning in large language models,'' in \emph{ICLR}, 2023.

\bibitem{zelikman2022star}
E.~Zelikman, J.~Mu, N.~D. Goodman, and Y.~T. Wu, ``Star: Self-taught reasoner bootstrapping reasoning with reasoning,'' in \emph{NeurIPS}, 2022.

\bibitem{yao2023tree}
S.~Yao, D.~Yu, J.~Zhao, I.~Shafran, T.~L. Griffiths, Y.~Cao, and K.~Narasimhan, ``Tree of thoughts: Deliberate problem solving with large language models,'' \emph{arXiv preprint arXiv:2305.10601}, 2023.

\bibitem{long2023large}
J.~Long, ``Large language model guided tree-of-thought,'' \emph{arXiv preprint arXiv:2305.08291}, 2023.

\bibitem{alpaca}
R.~Taori, I.~Gulrajani, T.~Zhang, Y.~Dubois, X.~Li, C.~Guestrin, P.~Liang, and T.~B. Hashimoto, ``Stanford alpaca: An instruction-following llama model,'' \url{https://github.com/tatsu-lab/stanford\_alpaca}, 2023.

\bibitem{gao2023llama}
P.~Gao, J.~Han, R.~Zhang, Z.~Lin, S.~Geng, A.~Zhou, W.~Zhang, P.~Lu, C.~He, X.~Yue \emph{et~al.}, ``Llama-adapter v2: Parameter-efficient visual instruction model,'' \emph{arXiv preprint arXiv:2304.15010}, 2023.

\bibitem{yu2021learning}
J.~{Yu}, X.~{Jiang}, Z.~{Qin}, W.~{Zhang}, Y.~{Hu}, and Q.~{Wu}, ``Learning dual encoding model for adaptive visual understanding in visual dialogue,'' \emph{IEEE TIP}, 2021.

\bibitem{hao2020towards}
W.~{Hao}, C.~{Li}, X.~{Li}, L.~{Carin}, and J.~{Gao}, ``Towards learning a generic agent for vision-and-language navigation via pre-training,'' in \emph{CVPR}, 2020.

\bibitem{radford2021learning}
A.~{Radford}, J.~W. {Kim}, C.~{Hallacy}, A.~{Ramesh}, G.~{Goh}, S.~{Agarwal}, G.~{Sastry}, A.~{Askell}, P.~{Mishkin}, J.~{Clark}, G.~{Krueger}, and I.~{Sutskever}, ``Learning transferable visual models from natural language supervision,'' in \emph{ICML}, 2021.

\bibitem{ilharco2019general}
G.~{Ilharco}, V.~{Jain}, A.~{Ku}, E.~{Ie}, and J.~{Baldridge}, ``General evaluation for instruction conditioned navigation using dynamic time warping,'' \emph{ViGIL@NeurIPS}, 2019.

\bibitem{openai}
OpenAI, ``Introducing chatgpt,'' \url{https://openai.com/blog/chatgpt}, 2022.

\end{thebibliography}
	
	%

\begin{IEEEbiography}[{\includegraphics[width=1in,height=1.25in,clip,keepaspectratio]{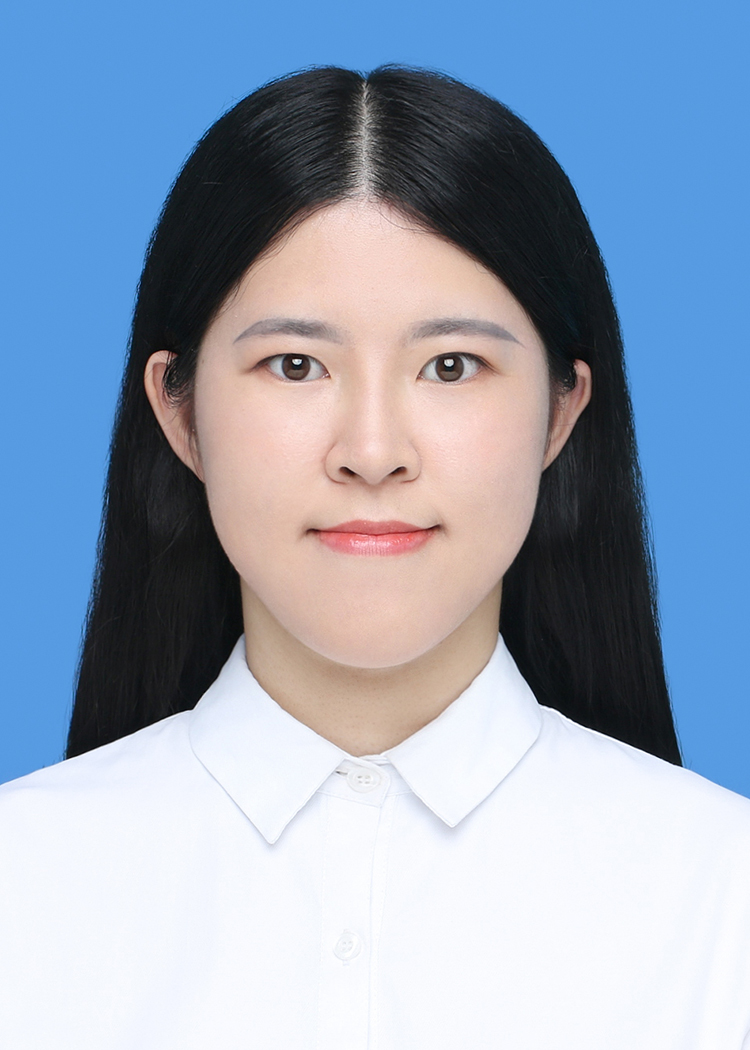}}]
{Bingqian Lin} is currently a postdoc researcher at Shanghai Jiao Tong University, working with Prof. Cewu Lu. She received her PhD degree from Sun Yat-sen University in 2024, advised by Prof. Xiaodan Liang. She received the B.E. and the M.E.
degree in Computer Science from University of
Electronic Science and Technology of China and
Xiamen University, in 2016 and 2019, respectively.
 Her research interests include vision-and-language understanding and embodied AI.
\end{IEEEbiography}

\begin{IEEEbiography}[{\includegraphics[width=1in,height=1.25in,clip,keepaspectratio]{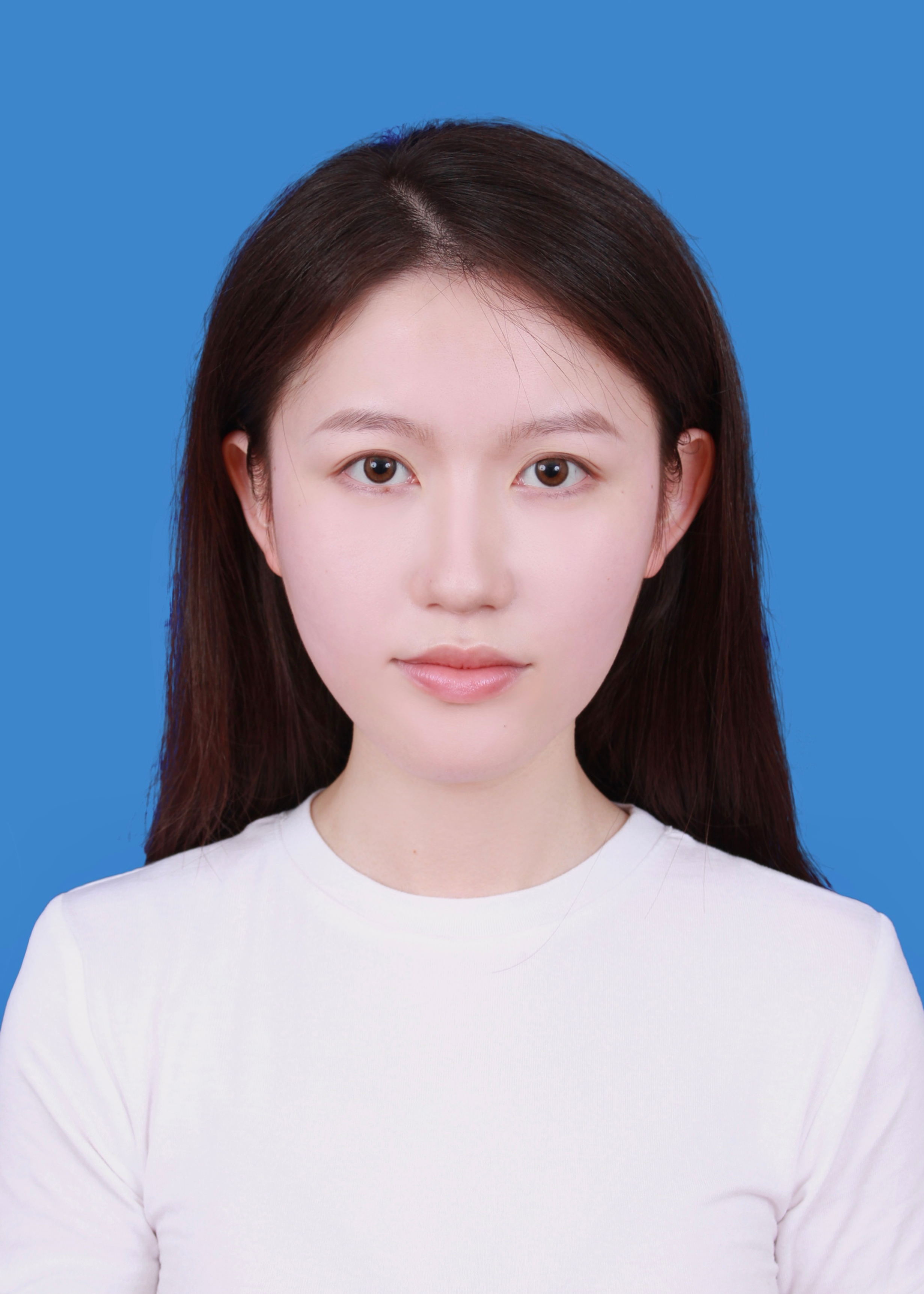}}]{Yunshuang Nie} received the B.E. degree in Sun Yat-sen University, Shenzhen, China, in 2023. She is currently working toward the M.E. in the school of intelligent systems engineering of Sun Yat-sen University. Her current research interest is vision-and-language understanding.
\end{IEEEbiography}

\begin{IEEEbiography}[{\includegraphics[width=1in,height=1.25in,clip,keepaspectratio]{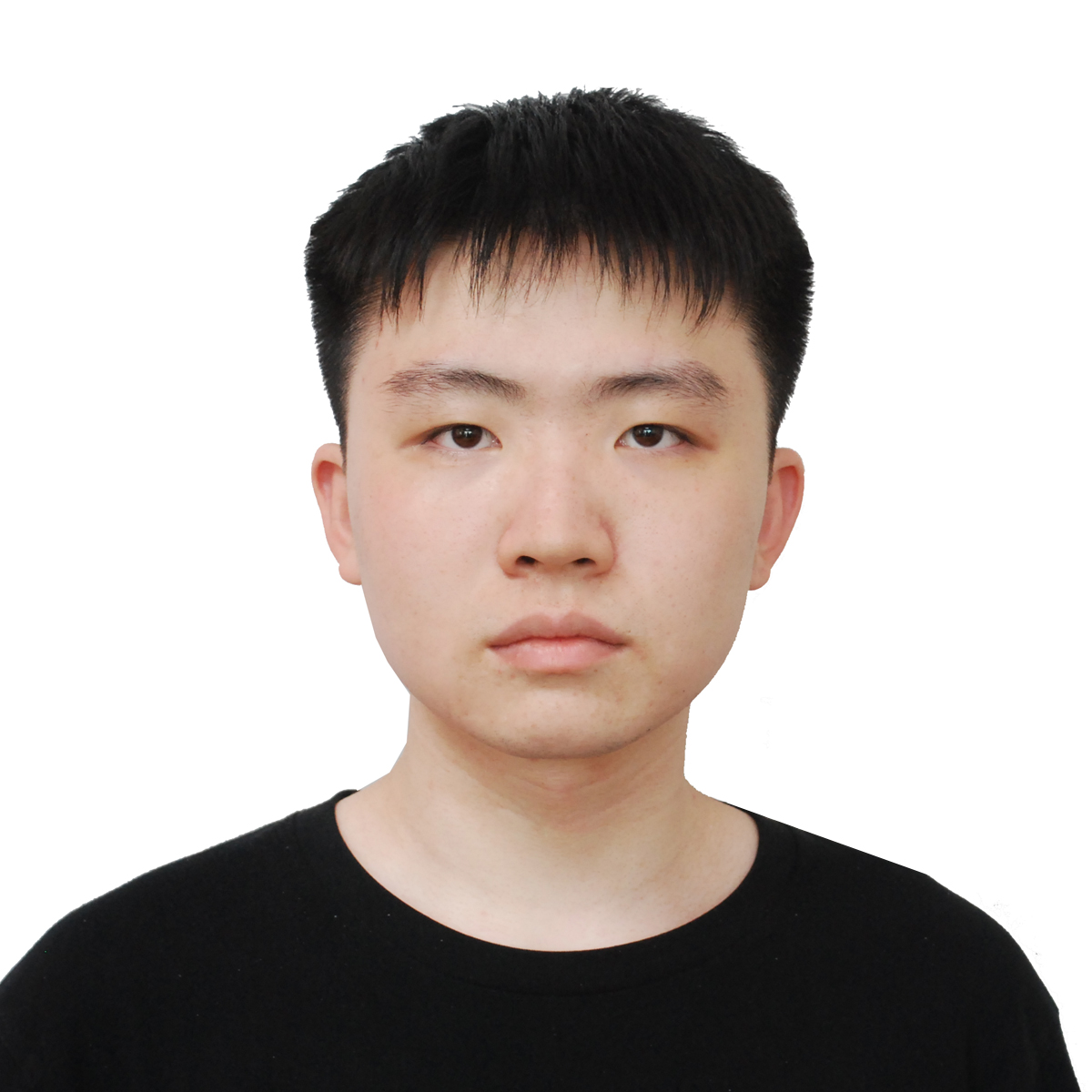}}]{Ziming Wei} is currently an undergraduate in the
school of intelligent systems engineering of Sun Yat-sen University. His current research interests include vision-and-language understanding, multimodality and embodied agent.
\end{IEEEbiography}

\begin{IEEEbiography}[{\includegraphics[width=1in,height=1.25in,clip,keepaspectratio]{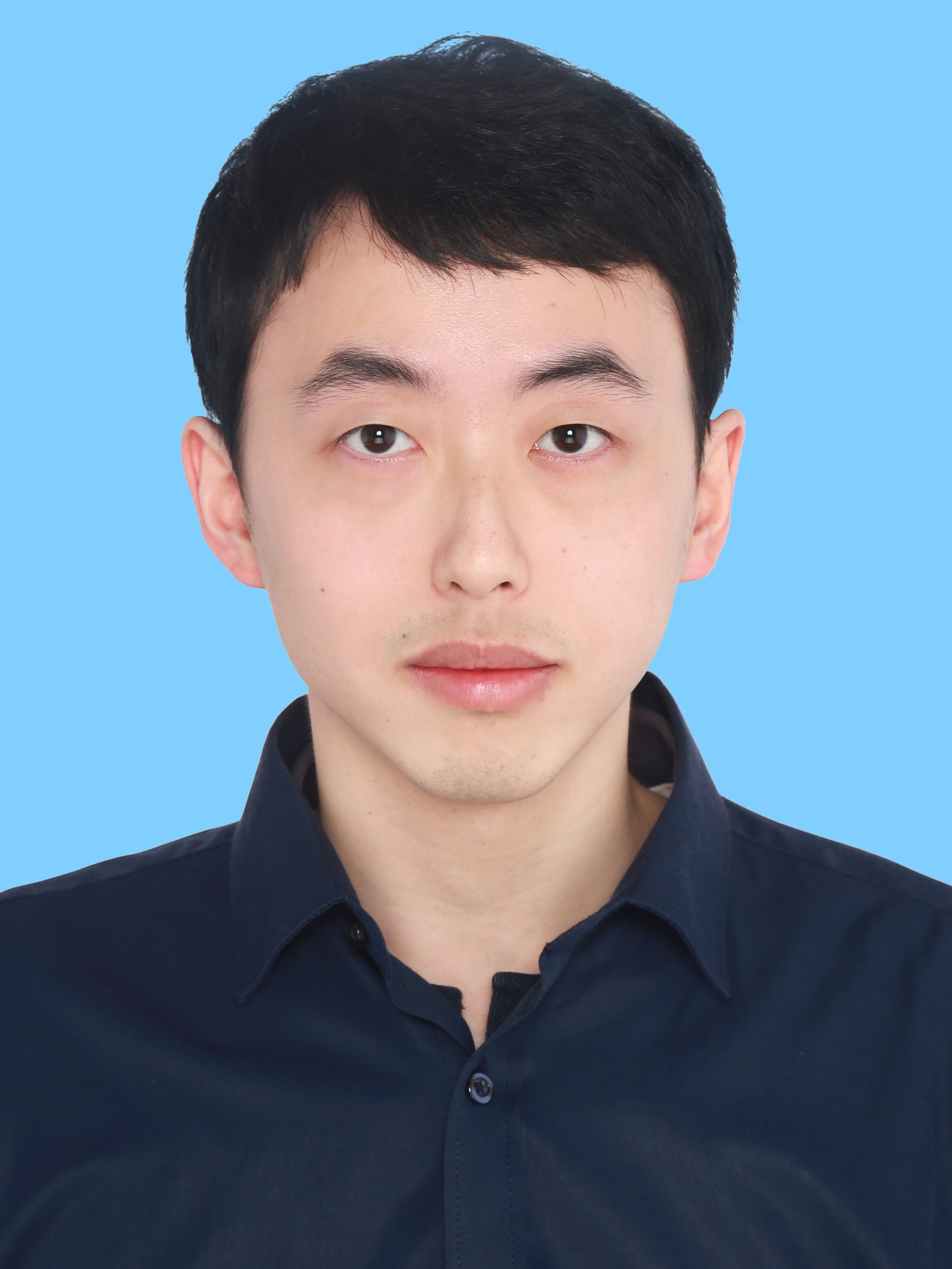}}]{Jiaqi Chen} received the B.E. and M.E. degrees from Xidian University and Sun Yat-sen University, in 2019 and 2021, respectively. He is currently working toward the Ph.D. degree in the Computer Science Department at The University of Hong Kong. His research interests include multi-modal reasoning and emboddied AI.
\end{IEEEbiography}

\begin{IEEEbiography}
[{\includegraphics[width=1in,height=1.25in, clip,keepaspectratio]{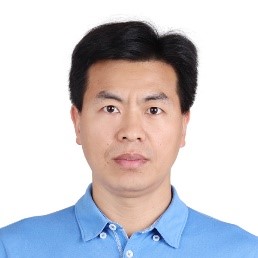}}]{Shikui Ma} received the M.S. degree in Northern Jiaotong University in 2003. He has been serving as an assistant vice president of Dataa Robotics company (https://www.dataarobotics.com/en) since 2015, where he is leading HARIX and AI R\&D team to consistently enhance their HARIX intelligent system, especially significantly improve the smart vision and motion capabilities of their robots via real-time digital twin, multi-modal fusion perception and advanced cognition, and deep reinforcement learning technologies. He has approximately 19 years of experience in communications technology industry, expertized in large-scale carrier-grade applications, especially competitive in Robotics, AI, Cloud, OSS and IPTV fields.
\end{IEEEbiography}

\begin{IEEEbiography}
[{\includegraphics[width=1in,height=1.25in, clip,keepaspectratio]{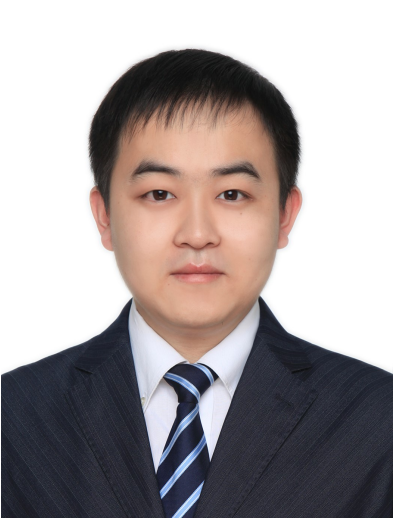}}]{Jianhua Han} received the Bachelor Degree in 2016 and Master Degree in 2019 from Shanghai Jiao Tong University, China. He is currently a re-searcher with the Noahs Ark Laboratory, Huawei Technologies Co ., Ltd. His research interests lie primarily in deep learning and computer vision.
\end{IEEEbiography}

\begin{IEEEbiography}
[{\includegraphics[width=1in,height=1.25in, clip,keepaspectratio]{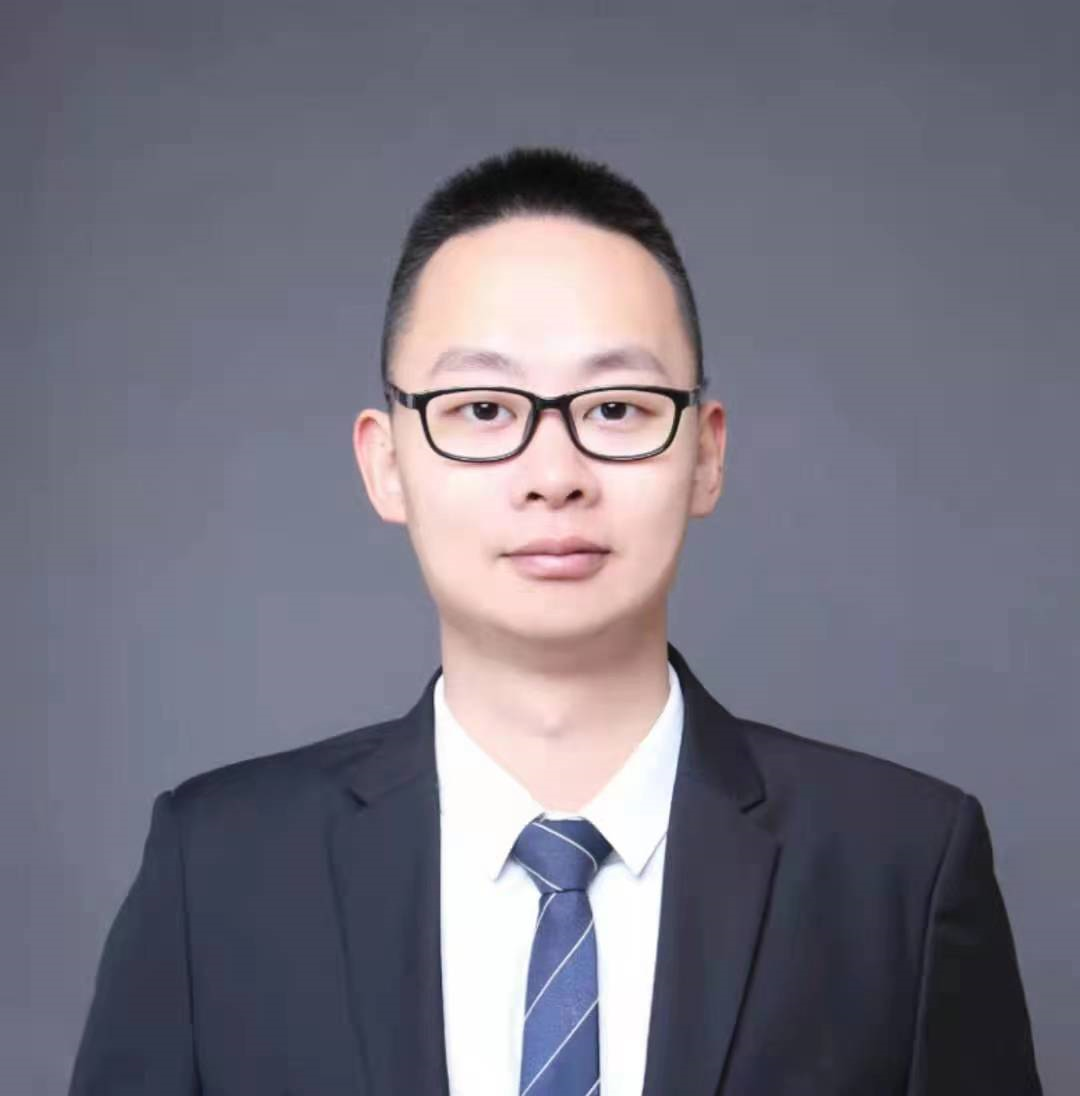}}]{Hang Xu} is currently a senior researcher in Huawei Noah Ark Lab. He received his BSc in Fudan University and Ph.D in Hong Kong University in Statistics. His research Interest includes multi-modality learning, machine Learning, object detection, and AutoML. He has published 70+ papers in Top AI conferences: NeurIPS, CVPR, ICCV, AAAI and some statistics journals, e.g. CSDA, Statistical Computing.
\end{IEEEbiography}

\begin{IEEEbiography}[{\includegraphics[width=1in,height=1.25in,clip,keepaspectratio]{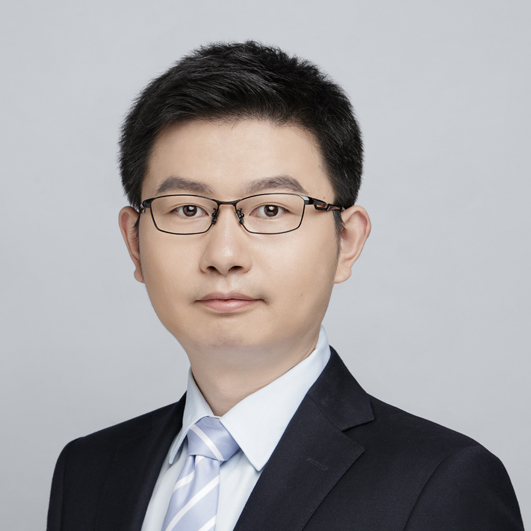}}]{Xiaojun Chang} (Senior Member, IEEE)
is a Professor at Australian Artificial Intelligence Institute, University of Technology Sydney. He is also an Honorary Professor at the School of Computing Technologies, RMIT University. Before joining UTS, he was an Associate Professor at the School of Computing Technologies, RMIT University, Australia. After graduation, he subsequently worked as a Postdoc Research Fellow at School of Computer Science, Carnegie Mellon University, Lecturer and Senior Lecturer in the Faculty of Information Technology, Monash University, Australia. He has spent most of his time working on exploring multiple signals (visual, acoustic, textual) for automatic content analysis in unconstrained or surveillance videos. He has achieved top performances in various international competitions, such as TRECVID MED, TRECVID SIN, and TRECVID AVS.
\end{IEEEbiography}

\begin{IEEEbiography}[{\includegraphics[width=1in,height=1.25in,clip,keepaspectratio]{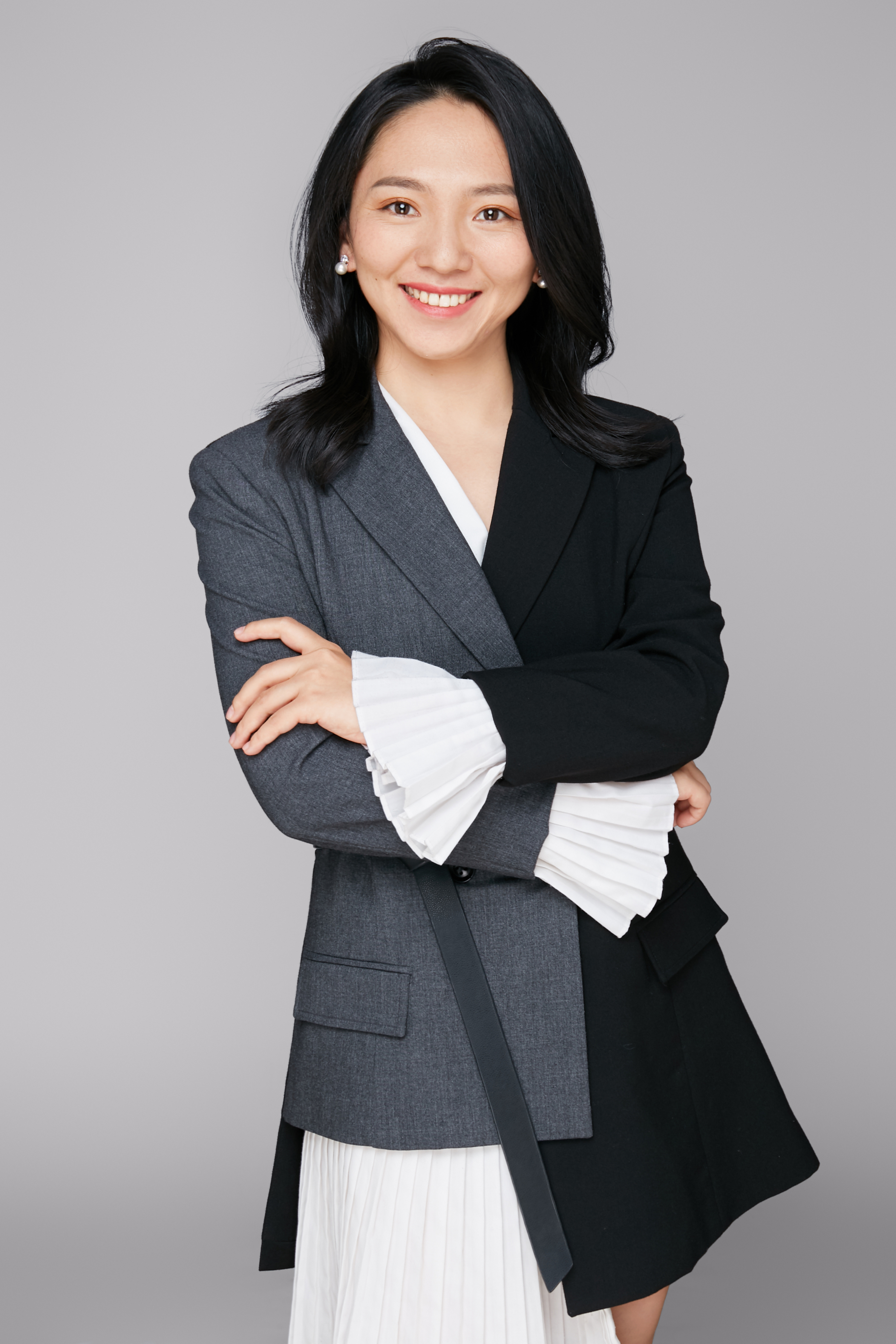}}]{Xiaodan Liang} is currently an Associate Professor at Sun Yat-sen University. She was a postdoc researcher in the machine learning department at Carnegie Mellon University, working with Prof. Eric Xing, from 2016 to 2018. She received her PhD degree from Sun Yat-sen University in 2016, advised by Liang Lin. She has published several cutting-edge projects on human-related analysis, including human parsing, pedestrian detection and instance segmentation, 2D/3D human pose estimation, and activity recognition.
\end{IEEEbiography}

	
	
	
	
	
	

\end{document}